\documentclass[abstract=true,paper=letter]{scrartcl}
\usepackage[margin=2.5cm]{geometry}

\usepackage{lineno,hyperref}
\usepackage[normal, bf]{caption}  
\usepackage{natbib}

\usepackage[english]{babel}
\usepackage[utf8x]{inputenc}
\usepackage[T1]{fontenc}
\usepackage[misc]{ifsym}

\usepackage{amsmath}
\usepackage{graphicx}
\usepackage{subfig}

\usepackage{booktabs}
\usepackage{algorithmicx}
\usepackage{algorithm}
\usepackage{algpseudocode}

\usepackage{tikz}
\usetikzlibrary{arrows}
\usepackage{color}
\usepackage{multirow}
\usepackage[T1]{fontenc}
\newcommand{\code}[1]{\texttt{#1}}

\begin{document}

\title{Optimising Optimisers with Push GP}

\author{Michael A. Lones\footnote{School of Mathematical and Computer Sciences, Heriot-Watt University, Edinburgh, UK, \href{mailto:m.lones@hw.ac.uk}{m.lones@hw.ac.uk}.}}

\date{}
\maketitle

\begin{abstract}
This work uses Push GP to automatically design both local and population-based optimisers for continuous-valued problems. The optimisers are trained on a single function optimisation landscape, using random transformations to discourage overfitting. They are then tested for generality on larger versions of the same problem, and on other continuous-valued problems. In most cases, the optimisers generalise well to the larger problems. Surprisingly, some of them also generalise very well to previously unseen problems, outperforming existing general purpose optimisers such as CMA-ES. Analysis of the behaviour of the evolved optimisers indicates a range of interesting optimisation strategies that are not found within conventional optimisers, suggesting that this approach could be useful for discovering novel and effective forms of optimisation in an automated manner.
\end{abstract}

%-------------------------------------------------
\section{Introduction}

This work is motivated by two issues. First, due to the innate constraints and biases of human thought, it is likely that manual design of optimisers only explores a subspace of optimiser design space. It is unlikely that this subspace contains optimal optimisers for all optimisation problems. Second, recent attempts to create novel optimisers from models of natural systems have been largely unsuccessful in broadening the scope of optimiser designs, instead tending only to generate variants of existing metaheuristic frameworks \citep{sorensen2015metaheuristics,lones2019mitigating}.

This work attempts to address both of these issues by using Genetic Programming (GP) to explore the broader space of optimisation algorithms, with the aim of discovering novel optimisation behaviours that differ from those used by existing algorithms. In order to make the optimiser search space as broad as possible, a Turing-complete language, \textit{Push}, is used to represent the optimisers, and the Push GP system is used to optimise them \citep{spector2001autoconstructive}. In \citep{lones2019instruction}, this approach was used to evolve local optimisers that can solve continuous-valued problems. In this work, this approach is extended to the population-based case, using Push GP to automatically design both local and population-based optimisers from primitive instructions.

The paper is organised as follows. Section \ref{related} reviews existing work on the automated design of optimisers. Section \ref{push} gives a brief overview of the Push language and the Push GP system, section \ref{evolving} describes how Push GP has been modified to support the evolution of population-based optimisers, and section \ref{evaluation} outlines how the optimisers are evaluated. Section \ref{results} presents results and analysis. Section \ref{conclusions} concludes.

\section{Related Work} \label{related}

There is a significant history of using GP to optimise optimisers. This can be divided into two areas: using GP to optimise GP, and using GP to optimise other kinds of optimiser. The former approaches use a GP system to optimise the solution generation operators of a GP framework  \citep{edmonds1998metagp,kantschik1999meta,spector2001autoconstructive}. Autoconstructive evolution \citep{spector2001autoconstructive} is a particularly open-ended approach to doing this in which programs contains code that generates their own offspring; also notable is that, like our work, it uses the Push language.

However, more relevant is previous work on using GP to optimise non-GP optimisers. Much of this work has taken place within the context of \textit{hyperheuristics}, which involves specialising existing optimisation frameworks so that they are better suited to solving particular problem classes. In this context, GP has been used to re-design components of evolutionary algorithms, such as their mutation \citep{woodward2012automatic}, recombination \citep{goldman2011self} and selection operators \citep{richter2018automated}, with the aim of making them better adapted to particular solution landscapes. Other hyperheuristic approaches have used GP to generate new optimisation algorithms by recombining the high-level building blocks of existing metaheuristic frameworks \citep{oltean2005evolving,martin2013evolving,ryser2016iterative,bogdanova2019franken}. Recently, this kind of approach has also been used to explore the design space of swarm algorithms, using grammatical evolution to combine high-level building blocks derived from existing metaheuristics \citep{bogdanova2019franken}. Our approach differs from this, and previous work in hyperheuristics, in that it focuses on designing optimisers largely from scratch. By not reusing or building upon components of existing optimisers, the intention is to reduce the amount of bias in the exploration of optimiser design space, potentially allowing the exploration of previously unexplored areas.

Another recent development, which has some similarities to our work, is the use of deep learning to optimise optimisers \citep{andrychowicz2016learning,wichrowska2017learned,metz2019learned}. So far these approaches have focused on improving the training algorithms used by deep learners, i.e. they are somewhat akin to using GP to optimise GP, though it is plausible that deep learning could be applied to the task of designing optimisers for non-neural domains. However, this is arguably an area in which GP is better suited than deep learning, since the optimisers produced by GP are likely to be far more efficient than those produced by deep learning. Efficiency is an important consideration for optimisers, since the same code is typically called over and over again during the course of an optimisation trajectory. Another advantage of GP is the relative interpretability of its solutions when compared to deep learning, and the potential that more general insights could be made into the design of optimisers by studying the code of evolved solutions.

\section{Methods} \label{methods}

\subsection{Push and Push GP} \label{push}

In this work, optimisation behaviours are expressed using the Push language. Push was designed for use within a GP context, and has a number of features that promote evolvability \citep{spector2001autoconstructive,spector2002genetic,spector2004push}. This includes the use of stacks, a mechanism that enables evolving programs to maintain state with less fragility than using conventional indexed memory instructions \citep{langdon2012genetic}. However, it is also Turing-complete, meaning that it is more expressive that many languages used within GP systems. Another notable strength is its type system, which is designed so that all randomly generated programs are syntactically valid, meaning that (unlike type systems introduced to more conventional forms of GP) there is no need to complicate the variation operators or penalise/repair invalid solutions. This is implemented by means of multiple stacks; each stack contains values of a particular type, and all instructions are typed, and will only execute when values are present on their corresponding type stacks. There are stacks for primitive data types (booleans, floats, integers) and each of these have both special-purpose instructions (e.g. arithmetic instructions for the integer and float stacks, logic operators for the boolean stack) and general-purpose stack instructions (push, pop, swap, duplicate, rot etc.) associated with them. Another important stack is the execution stack. At the start of execution, the instructions in a Push program are placed onto this stack and can be manipulated by special instructions; this allows behaviours like looping and conditional execution to be carried out. Finally, there is an input stack, which remains fixed during execution. This provides a way of passing non-volatile constants to a Push program; when popped from the input stack, corresponding values get pushed to the appropriate type stack. Push programs are evolved using the Push GP system. Since Push programs are basically a list of instructions, they can be represented as a linear array and manipulated using genetic algorithm-like mutation and crossover operators.

\subsection{Evolving Population-Based Optimisers} \label{evolving}

In order to evolve population-based optimisers, this work uses a modified version of \code{Psh} (\url{http://spiderland.org/Psh/}), a Java implementation of Push GP. To allow programs to store and manipulate search points, an extra vector type has been added to the Push language. This represents search points as fixed-length floating point vectors, and these can be manipulated using the special-purpose vector instructions shown in Table \ref{tab:vector}; see \citep{lones2019instruction} for more details about these instructions. Evolutionary parameters are shown in Table \ref{table:settings}.

\begin{table}[!tb]
	\centering
	\caption{Psh parameter settings}
	\vspace{1mm}
	\begin{tabular}{@{}p{0.98\textwidth}@{}}
		\toprule
		Population size  = 200\\
		Maximum generations = 50\\
		Tournament size = 5\\
		Program size limit = maximum of 100 instructions\\
		Execution limit = maximum of 100 instruction executions per move\\
		Instructions =
		\code{boolean/float/integer/vector.\{dup flush pop rand rot shove stackdepth swap yank yankdup\}; boolean.\{= and fromfloat frominteger not or xor\}; exec.\{= do*count do*range do*times if iflt noop\}; float.\{\% * + - / < = > abs cos erc exp fromboolean frominteger ln log max min neg pow sin tan\}; input.\{inall inallrev index\}; integer.\{\% * + - / < = > abs erc fromboolean fromfloat ln log max min neg pow\}; vector.\{* / + - apply between dim+ dim* dprod mag pop scale urand wrand zip\}; false; true}\\
		\bottomrule
	\end{tabular}
	\label{table:settings}
\end{table}

\begin{table}[htb!]
	\caption{Vector stack instructions}
	\vspace{1mm}
	\centering
	\label{tab:vector}
	\begin{tabular}{@{}llll@{}}
		\toprule
		Instruction&Pop from&Push to&Description\\
		\midrule
		\code{vector.+}&vector, vector&vector&Add two vectors\\
		\code{vector.-}&vector, vector&vector&Subtract two vectors\\
		\code{vector.*}&vector, vector&vector&Multiply two vectors\\
		\code{vector./}&vector, vector&vector&Divide two vectors\\
		\code{vector.scale}&vector, float&vector&Scale vector by scalar\\
		\code{vector.dprod}&vector, vector&float&Dot product of two vectors\\
		\code{vector.mag}&vector&float&Magnitude of vector\\
		\code{vector.dim+}&vector, float, int&vector&Add float to specified component\\
		\code{vector.dim*}&vector, float, int&vector&Multiple specified component by float\\
		\code{vector.apply}&vector, code&vector&Apply code to each component \\
		\code{vector.zip}&vector, vector, code&vector&Apply code to each pair of components\\
		\code{vector.between}&vector, vector, float&vector&Generate point between two vectors\\
		\midrule
		\code{vector.rand}&&vector&Generate random vector of floats\\
		\code{vector.urand}&&vector&Generate random unit vector\\
		\code{vector.wrand}&float&vector&Generate random vector within bounds\\
		\midrule
		\code{vector.current}&integer&vector&Get current point of given pop member\\
		\code{vector.best}&integer&vector&Get best point of given pop member\\
		\bottomrule
	\end{tabular}
\end{table}

Algorithm \ref{alg::evaluation} outlines the procedure for evaluating evolved Push optimisers. To reduce evolutionary effort, a Push program is only required to carry out a single move, or optimisation step, each time it is called. In order to generate an optimisation trajectory within a given search space, the Push program is then called multiple times by an outer loop until a specified evaluation budget has been reached. After each call, the value at the top of the Push program's vector stack is popped and the corresponding search point is evaluated. The objective value of this search point, as well as information about whether it was an improving move and whether it moved outside the problem's search bounds, are then passed back to the Push program via the relevant type stacks. Since the contents of a program's stacks are preserved between calls, Push programs have the capacity to build up their own internal state during the course of an optimisation run, and consequently the potential to carry out different types of moves as search progresses.

\begin{algorithm}[p!]
	\caption{Evaluating an evolved Push GP optimiser}
	\label{alg::evaluation}
	\begin{algorithmic}[1]
		\State $\mathit{fitness} \gets 0$
		\For{$\mathit{repeats}$} \Comment{Measure fitness over multiple optimisation runs}
		\State $\mathit{pbest} \gets \infty$
		\For{$p \gets 1,\mathit{popsize}$} \Comment{Initialise population state}
		\State $\mathit{prog}_p \gets$ copy of evolved Push program
		\State $\textsc{clearstacks}(\mathit{prog}_p)$
		\State $\mathit{point}_p \gets$ random initial point within search bounds
		\State $\mathit{value}_p \gets \textsc{evaluate}(\mathit{point}_p)$
		\State \textsc{push}($\mathit{point}_p$, $\mathit{prog}_p$.\code{vector}) \Comment{Pass initial search point to program}
		\State \textsc{push}($\mathit{value}_p$, $\mathit{prog}_p$.\code{float}) \Comment{Pass initial objective value to program}
		\State \textsc{push}(\code{true}, $\mathit{prog}_p$.\code{boolean})
		\State \textsc{push}(bounds, $\mathit{prog}_p$.\code{input}) \Comment{Put search space bounds on input stack}
		\State $\mathit{bestval}_p\gets \mathit{value}_p$
		\If{$\mathit{bestval}_p < \mathit{pbest}$}
		\State $\mathit{pbest} \gets \mathit{bestval}_p, \hspace{2mm} \mathit{pbestindex} \gets p$
		\EndIf
		\EndFor
		\For{$m \gets 1,\mathit{moves}$}
		\For{$p \gets 1,\mathit{popsize}$}
		\State \textsc{push}($m$, $\mathit{prog}_p$.\code{integer}) \Comment{Pass move number to program}
		\State \textsc{push}($p$, $\mathit{prog}_p$.\code{integer}) \Comment{Pass population index to program}
		\State \textsc{push}($pbestindex$, $\mathit{prog}_p$.\code{integer}) \Comment{Pass  index of pbest to program}
		\State $\mathit{previous} \gets \mathit{value}_p$
		\State $\textsc{execute}(\mathit{prog}_p)$
		\State $\mathit{point}_p \gets$ \textsc{peek}($\mathit{prog}_p$.\code{vector})
		\Comment{Get next search point from program}
		
		\If{$\mathit{point}_p$ is within search bounds}
		\State $\mathit{value}_p \gets \textsc{evaluate}(\mathit{point}_p)$
		\If{$\mathit{value}_p < \mathit{bestval}_p$}
		\State $\mathit{bestval}_p \gets \mathit{value}_p, \hspace{2mm} \mathit{best}_p \gets \mathit{point}_p$
		\EndIf
		\If{$\mathit{value}_p < \mathit{previous}$}
		\State \textsc{push}(\code{true}, $\mathit{prog}_p$.\code{boolean}) \Comment{Tell program it improved}
		\Else
		\State \textsc{push}(\code{false}, $\mathit{prog}_p$.\code{boolean}) \Comment{Tell program it didn't improve}
		\State \textsc{push}($\mathit{best}_p$, $\mathit{prog}_p$.\code{vector}) \Comment{and remind it of its best point}
		\EndIf
		\State \textsc{push}($\mathit{value}_p$, $\mathit{prog}_p$.\code{float}) \Comment{Pass new objective value}
		\Else
		\State \textsc{push}(\code{false}, $\mathit{prog}_p$.\code{boolean})
		\State \textsc{push}($\infty$, $\mathit{prog}_p$.\code{float}) \Comment{Or indicate move was out of bounds}
		\EndIf
		
		\State \algorithmicif \hspace{0.5mm} $\mathit{best}_p < \mathit{pbest}$ \hspace{0.2mm} \algorithmicthen \hspace{0.5mm} $\mathit{pbest} \gets \mathit{best}_p$
		\EndFor
		\EndFor
		\State $\mathit{fitness} \gets \mathit{fitness} + \mathit{pbest}$
		\EndFor
		\State $\mathit{fitness} \gets \mathit{fitness} / \mathit{repeats}$ \Comment{Mean of best objective values found in each repeat}
	\end{algorithmic}
\end{algorithm}

To handle population-based optimisation, multiple copies of the Push program are run in parallel, one for each population member. Each copy of the program has its own stacks, so population members are able to build up their internal state independently. Population members are persistent, meaning there is no explicit mechanism to create or destroy them during the course of an optimisation run. To allow coordination between population members, two extra instructions are provided, \code{vector.current} and \code{vector.best}. These both look up information about another population member's search state, returning either its current or best seen point of search. The target population member is determined by the value at the top of the integer stack (modulus the population size to ensure a valid number); if this stack is empty, or contains a negative value, the current or best search point of the current population member is returned. This sharing mechanism, combined with the use of persistent search processes, means that the evolved optimisers resemble swarm algorithms in their general mechanics. However, there is no selective pressure to use these mechanisms in any particular way, so evolved optimisers are not constrained by the design space of existing swarm optimisers.

\subsection{Evaluation} \label{evaluation}

Evolved optimisers are evaluated using a selection of functions taken from the widely used CEC 2005 real-valued parameter optimisation benchmarks. These are all minimisation problems, meaning that the aim is to find the input vector (i.e. the search point) that generates the lowest value when passed as an argument to the function. The functions used during fitness evaluation, which were selected to provide a diverse range of optimisation landscapes, are:

\begin{itemize}
	\item $F_1$, the sphere function, a separable unimodal bowl-shaped function. It is the simplest of the benchmarks, and can be solved by gradient descent.
	\item $F_9$, Rastrigin's function, has a large number of regularly spaced local optima whose magnitudes curve towards a bowl where the global minimum is found. The difficulty of this function lies in avoiding the many local optima on the path to the global optimum, though it is made easier by the regular spacing, since the distance between local optima basins can in principle be learnt.
	\item $F_{12}$, Schwefel's problem number 2.13, is multimodal and has a small number of peaks that can be followed down to a shared valley region. Gradient descent can be used to find the valley, but the difficulty lies in finding the global mimimum, since it contains multiple irregularly-spaced local optima.
	\item $F_{13}$ is a composition of Griewank's and Rosenbrock's functions. This composition leads to a complex surface that is highly multimodal and irregular, and hence challenging for optimisers to navigate.
	\item $F_{14}$, a version of Schaffer's $F_6$ Function, comprises concentric elliptical ridges. In the centre is a region of greater complexity where the global optimum lies. It is challenging due to the lack of useful gradient information in most places, and the large number of local optima.
\end{itemize}

\noindent To discourage overfitting to a particular problem instance, random transformations are applied to each dimension of these functions when they are used to measure fitness during the course of an evolutionary run. Random translations (of up to $\pm$50\% for each axis) prevent the evolving optimisers from learning the location of the optimum, random scalings (50-200\% for each axis) prevent them from learning the distance between features of the landscape, and random axis flips (with 50\% probability per axis) prevent directional biases, e.g. learning which corner of the landscape contains the global optimum. Fitness is the mean of 10 optimisation runs, each with random initial locations and random transformations. The 10-dimensional versions of the problems are used for training, with an evaluation budget of 1E+3 fitness evaluations (FEs). For the results tables and figures shown in the following section, the best-of-run optimisers are reevaluated over the CEC 2005 benchmark standard of 25 optimisation runs, and random transformations are not applied.

\section{Results} \label{results}

\begin{figure}[p!]
	\centering
	\includegraphics[width=0.85\columnwidth]{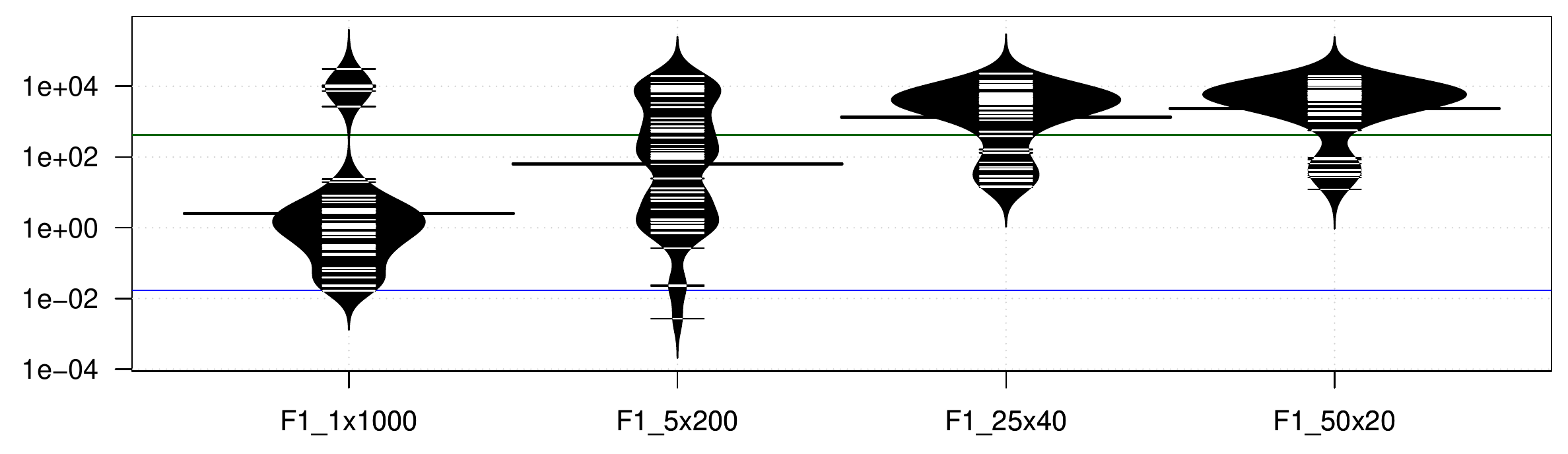}
	\includegraphics[width=0.85\columnwidth]{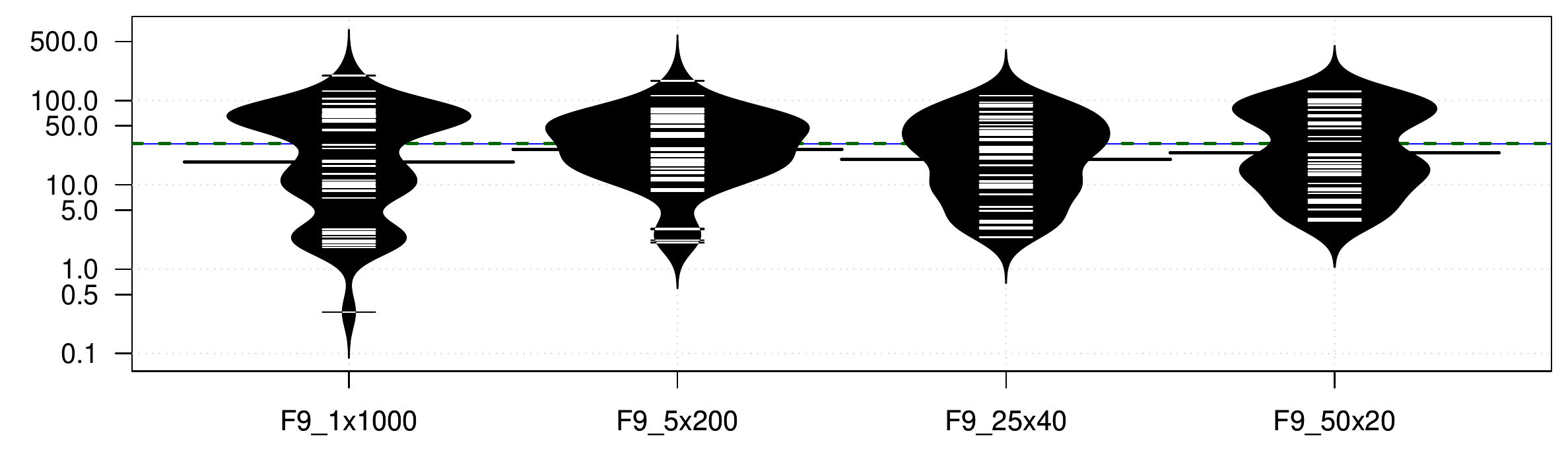}
	\includegraphics[width=0.85\columnwidth]{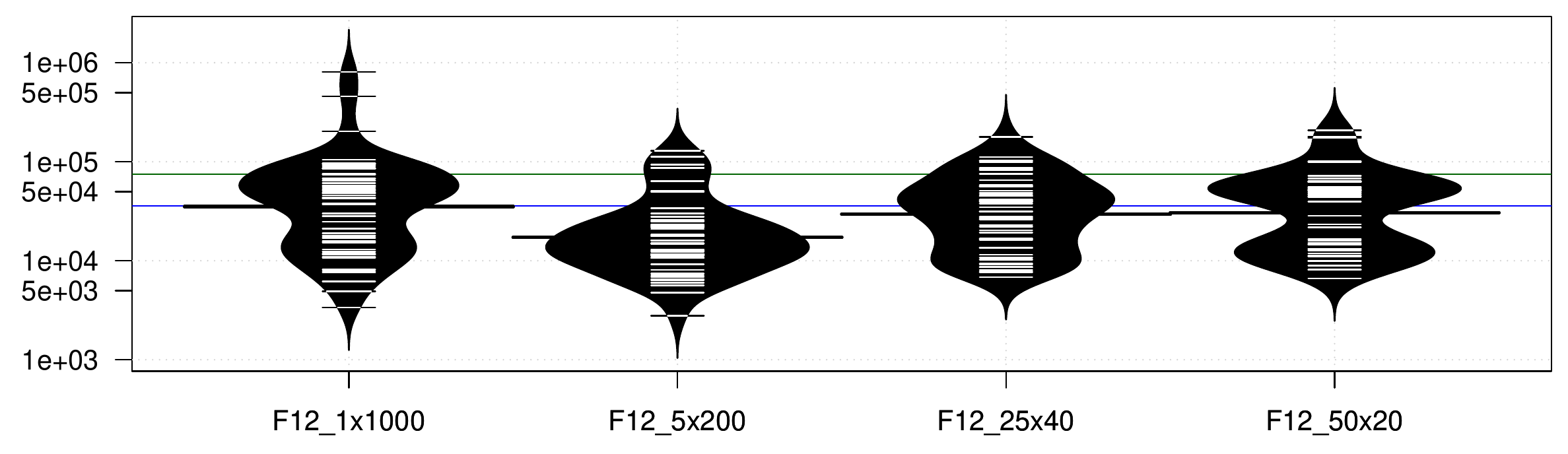}
	\includegraphics[width=0.85\columnwidth]{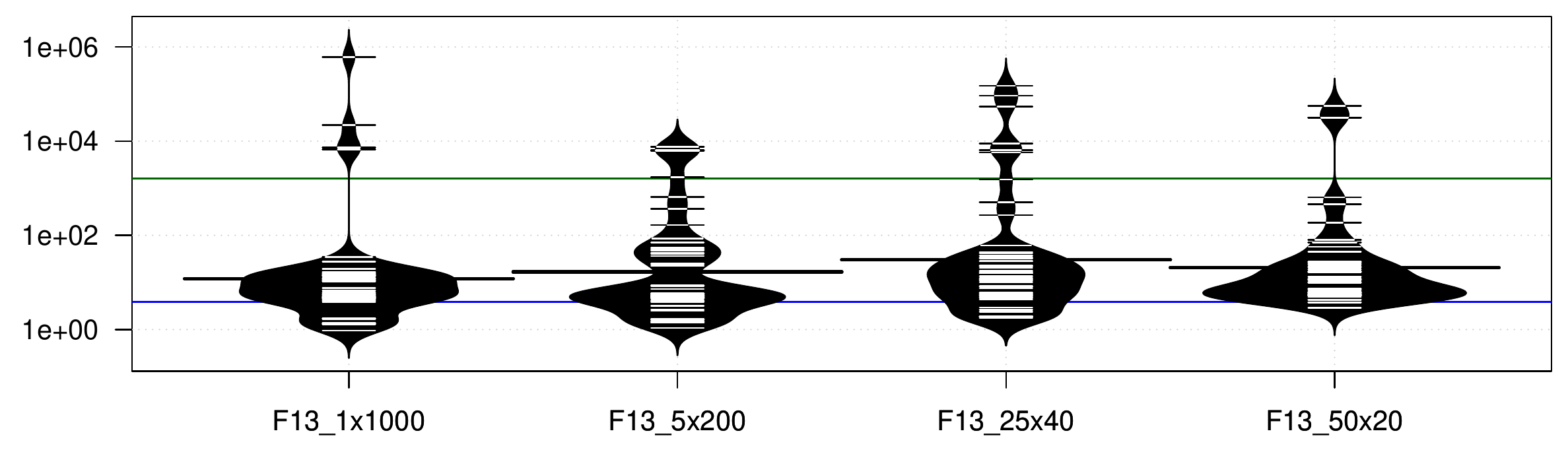}
	\includegraphics[width=0.85\columnwidth]{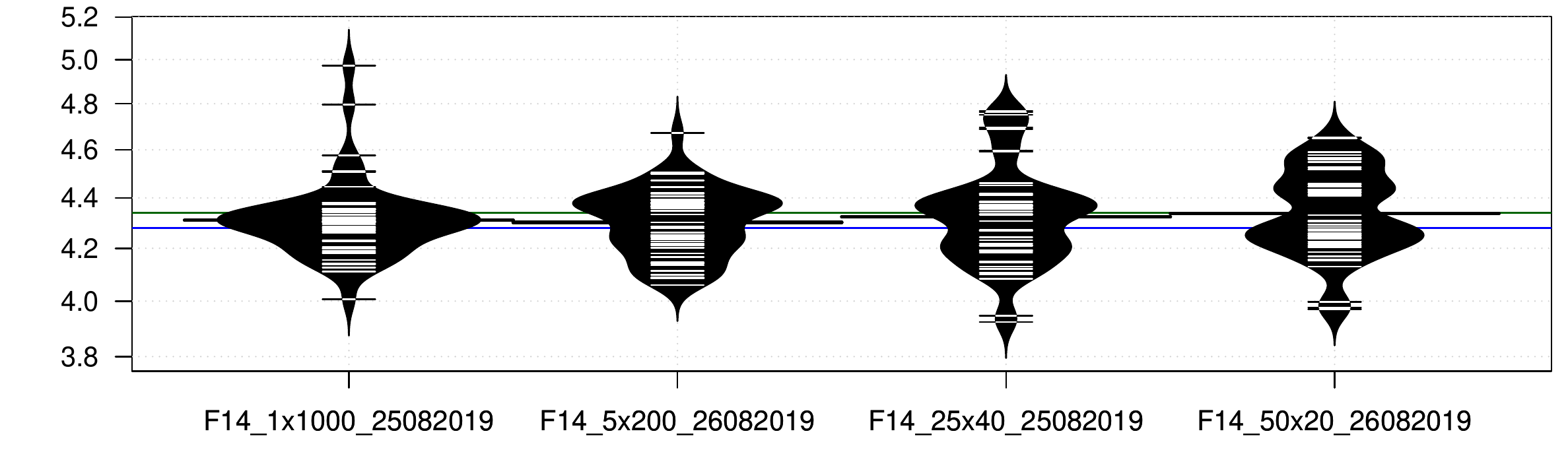}
	\caption{Fitness distributions across 50 runs for each problem. The value shown for each each run is the mean fitness of the best solution over 25 reevaluations. Published results for CMA-ES (blue) and DE (green) are also shown.}
	\label{fig:distributions}
\end{figure}

For a population-based optimiser, the 1E+3 evaluation budget can be split between the population size and the number of iterations/generations in different ways. In these experiments, splits of (population size $\times$ iterations) 50$\times$20, 25$\times$40, 5$\times$200 and 1$\times$1000 are used. The latter is included to give a comparison against local search, i.e. optimisers which only use a single point of search. Fig. \ref{fig:distributions} shows the fitness distributions over 50 evolutionary runs for each of these configurations, where fitness is the mean error when the best-of-run optimisers are reevaluated over 25 optimisation runs. To give an idea of how these error rates compare to established general purpose optimisers, Fig. \ref{fig:distributions}  also reproduces the mean errors achieved by two algorithms from the original CEC 2005 competition. G-CMA-ES \citep{auger2005restart} is a variant of the Covariance Matrix Adaptation Evolution Strategy (CMA-ES) with the addition of restarts and an increasing population size at each restart; it is a relatively complex algorithm and is generally regarded as the overall winner of the CEC 2005 competition. Differential Evolution (DE) \citep{ronkkonen2005real}, although less successful than G-CMA-ES in the competition, is another example of a well-regarded population-based optimiser.

Fig. \ref{fig:distributions} compares the ability of Push GP to find optimisers with different trade-offs between population size and number of iterations. The distributions show that this trade-off is more important for some problems than others. For $F_1$, better optimisers are generally found for smaller population sizes, with the 1$\times$1000 distribution having the lowest mean error. This makes sense, because the unimodal $F_1$ landscape favours intensification over diversification. For $F_{12}$, the sweet spot appears to be for 5$\times$200, possibly reflecting the number of peaks in the landscape, i.e. 5.  For the other problems, the differences appear relatively minor, and effective optimisers could be evolved for all configurations. In most cases, the best optimiser for a particular problem is an outlier within the distributions, so may not reflect any intrinsic benefit of one configuration over another. That said, four of these best-in-problem classifiers used small populations (2 with 1$\times$1000 and 2 with 5$\times$200), so maybe it is easier to find effective optimisers that use small populations than larger ones. 

Perhaps more importantly, Fig. \ref{fig:distributions} shows that the Push GP runs found at least one optimiser that performed better, on average, than CMA-ES and DE. For the simplest problem $F_1$, there was only one evolved optimiser that beat the general purpose optimisers. For the other problems, many optimisers were found that performed better. This reflects the results in \citep{lones2019instruction}, and is perhaps unsurprising given that the capacity to overfit problems is a central motivation for existing work in hyperheuristics. However, an important difference in this paper is the use of random problem transformations during training, since this causes the problems to exhibit greater generality, preventing optimisers from over-learning specific features of the landscape. The results suggest that this does not effect the ability of evolved optimisers to out-perform general purpose optimisers.

\begin{table*}[tb!]
	\caption{Generality of evolved optimisers. For each optimiser, mean errors are shown for 25 optimisation runs on 10D and 30D problems. The mean rank including (and excluding) the problem the optimiser was trained on is also shown, and the best result for each combination of problem dimensionality (D) and fitness evaluation budget (FEs) is underlined for each problem number and ranking.}\label{table:errors}
	\centering
	\vspace{2mm}
	\begin{tabular}{@{}lllllllll@{}}
		\toprule
		D & FEs & Optimiser & $F_1$ & $F_9$ & $F_{12}$ & $F_{13}$ & $F_{14}$ & $\overline{\mathrm{Rank}}$\\
		\midrule
		10 & 1E+3 & CMA-ES & 1.70E$-$2 & 3.07E+1 & 3.59E+4 & 3.84E+0 & 4.28E+0 & \underline{3.4}\\
		& & DE & 4.21E+2 & 3.11E+1 & 7.48E+4 & 1.62E+3 & 4.34E+0 &5.0\\
		& & $F_{1}$ best & \underline{2.48E$-$3}&7.28E+1&3.29E+4&5.26E+0&4.47E+0&4.0 (4.8)\\
		& &$F_{19}$ best & 1.32E+4&\underline{3.27E$-$1}&9.32E+3&1.18E+0&4.86E+0&3.6 (4.3)\\
		& & $F_{12}$ best & 3.10E+3&	7.28E+0&\underline{2.79E+3}&2.43E+0&4.52E+0&\underline{3.4} (4.0)\\
		& & $F_{13}$ best & 3.56E+4&2.44E+0&4.63E+4&\underline{1.05E+0}&4.82E+0&4.2 (5.0)\\
		& & $F_{14}$ best & 4.11E+2&7.76E+1&9.97E+4&2.69E+2&\underline{4.04E+0}&4.4 (5.3)\\
		\cmidrule{2-9}
		& 1E+4 & CMA-ES & \underline{5.20E$-$9} & 6.21E+0 & 2.98E+3 & 9.71E$-$1 & 3.91E+0 &\underline{2.8}\\
		& & DE & 2.00E+1 & \underline{5.49E$-$9} & 1.64E+4 & 9.05E+0 & 4.02E+0 &4.2\\
		& & $F_{1}$ best & 2.44E$-$6&8.05E+1&2.36E+4&3.60E+0&4.50E+0&4.8 (5.5)\\
		& & $F_{9}$ best & 1.45E$-$3&2.06E$-$1&7.72E+3&7.04E$-$1&4.85E+0&3.8 (3.8)\\
		& & $F_{12}$ best & 5.96E$-$4&7.47E$-$2&\underline{3.93E+2}&4.98E$-$1&4.21E+0&3.0 (3.5)\\
		& & $F_{13}$ best & 1.51E$-$4&3.66E$-$6&3.07E+4&\underline{3.45E$-$1}&4.90E+0&4.0 (4.8)\\
		& & $F_{14}$ best & 1.37E+1&5.16E+1&3.77E+4&1.62E+1&\underline{3.57E+0}&5.4 (6.5)\\
		\cmidrule{1-9}
		30 & 1E+3 & CMA-ES & \underline{8.16E+2} & 2.53E+2 & 1.67E+6 & 1.14E+2 & 1.42E+1 &3.2\\
		& & DE & 2.06E+4 & 3.77E+2 & 1.53E+6 & 1.62E+5 & 1.41E+1 &4.2\\
		& & $F_{1}$ best & 7.75E+4&4.36E+2&1.07E+6&3.47E+4&1.45E+1&5.2 (5.0)\\
		& & $F_{9}$ best & 7.63E+4&3.24E+2&1.07E+6&4.00E+3&1.45E+1&4.2 (4.3)\\
		& & $F_{12}$ best & 5.74E+4&1.18E+2&3.46E+5&3.62E+1&1.44E+1&\underline{2.8} (3.0)\\
		& & $F_{13}$ best & 1.63E+5&\underline{1.00E+2}&\underline{1.73E+5}&\underline{1.84E+1}&1.47E+1&3.4 (4.0)\\
		& & $F_{14}$ best & 2.14E+4&4.15E+2&2.19E+6&3.52E+4&\underline{1.38E+1}&4.6 (5.5)\\
		\cmidrule{2-9}
		& 1E+4 & CMA-ES & \underline{5.42E$-$9} & 4.78E+1 & 2.51E+5 & 3.80E+0 & 1.38E+1 &\underline{2.2}\\
		& & DE & 4.71E+0 & 9.85E+1 & 9.29E+5 & 1.02E+2 & 1.39E+1 &4.0\\
		& & $F_{1}$ best & 1.36E+2&3.68E+2&4.08E+5&4.18E+1&1.44E+1&4.8 (5.0)\\
		& & $F_{9}$ best & 6.40E+4&3.27E+2&1.09E+6&3.52E+3&1.46E+1&6.0 (6.3)\\
		& & $F_{12}$ best & 5.97E$-$2&5.76E+0&\underline{3.43E+4}&5.00E+0&1.41E+1&2.4 (2.8)\\
		& & $F_{13}$ best & 2.44E+4&\underline{5.04E$-$2}&1.26E+5&\underline{1.42E+0}&1.47E+1&3.4 (4.0)\\
		& & $F_{14}$ best & 1.64E+2&3.33E+2&1.17E+6&3.97E+3&\underline{1.33E+1}&5.2 (6.3)\\
		\bottomrule
	\end{tabular}
\end{table*}

The ability to out-perform general purpose optimisers on the problem on which they were trained is arguably not that important. Of more interest is how they generalise to larger and different problems. Table \ref{table:errors} gives an insight into this, showing how well the best evolved optimiser for each training problem generalises to larger instances of the same problem and to the other four problems. Mean error rates are shown both for the 10-dimensional problems with the 1E+3 evaluation budget used in training, and for 30-dimensional versions of the same problems and 1E+4 evaluation budgets. First of all, these figures show that the evolved optimisers do not stop progressing after the 1E+3 solution evaluations on which they were trained, since they make significantly more progress on the same problem when given a budget of 1E+4 solution evaluations. Also, it is evident that most of the optimisers generalise well to 30-dimensional versions of the same problem. The best optimisers evolved on the 10D $F_{12}$, $F_{13}$ and $F_{14}$ problems do particularly well in this regard, outperforming CMA-ES and DE on both the 10D and 30D versions of the problems. The $F_1$ optimiser is the only one which generalises relatively poorly, being beaten by CMA-ES, DE and several of the other optimisers on the 30D version.

The most interesting insight from Table \ref{table:errors} is that many of the optimisers also generalise to other problems. For the 10D, 1E+3 evaluations case, all of the optimisers do better than DE when their average rank is taken across all five problems. More surprisingly, the $F_{12}$ optimiser does as well as CMA-ES across all problems, despite only having been trained on one of them. Its average rank does drop slightly when its $F_{12}$ rank is removed from the calculation of its average rank, suggesting it doesn't generalise quite as well as CMA-ES on the 10D problems. However, the figures for the 30D case are even more surprising, with the $F_{12}$ optimiser doing better across the five problems (even with $F_{12}$ discounted) than CMA-ES. Also notable is that the $F_{13}$ optimiser comes first in three out of the five 30D problems, though this is balanced by coming last in the other two. CMA-ES does do slightly better than the $F_{12}$ optimiser when given a budget of 1E+4 solution evaluations, but the difference is slight, and the best mean error rates for the four most difficult problems are found by the evolved optimisers.

\begin{table*}[bt!]
	\caption{Evolved Push expressions of best-in-problem optimisers}
	\label{tab:optimisers}
	\vspace{1mm}
	\centering
	\begin{tabular}{@{}lp{14.5cm}@{}}
		\toprule
		$F_1$&\code{(exec.dup float.- vector.- float.pop vector.zip vector.zip integer.swap float.cos float.- float.cos float.- float.yank vector.best vector.wrand float.abs float.dup float.frominteger vector.- vector.dim*)}\\
		$F_9$&\code{(input.stackdepth float.frominteger vector.yank vector.wrand boolean.dup integer.fromboolean vector.swap integer.rot float.frominteger float.sin vector.yank vector.shove vector.dim+ vector.yank 0.0 float.> input.inall boolean.not 1 boolean.dup vector.pop boolean.stackdepth)}\\
		$F_{12}$&\code{(vector.stackdepth vector.swap float.fromboolean integer.fromboolean integer.rand vector.dim+ float.+ vector.swap integer.rand 0 vector.swap integer.max integer.= vector.stackdepth integer.dup vector.- integer.dup integer.rand vector.-  vector.dim+ vector.mag float.frominteger float.tan integer.rot vector.dim+)}\\
		$F_{13}$&\code{(integer.- float.sin vector.wrand integer.yankdup vector.dim* vector.- input.inall float.sin vector.-)}\\
		$F_{14}$&\code{(float.< float./ vector.best vector.yankdup float.ln float.max float.stackdepth 0.48999998 float.abs vector.between vector.wrand vector.scale integer.yank input.index vector.- float.rand float.neg 0.97999996 float.- 0.97999996 vector.wrand vector.scale vector.-)}\\
		\bottomrule
	\end{tabular}
\end{table*}

% trim={<left> <lower> <right> <upper>}
\begin{figure}[tb!]
	\centering
	\includegraphics[width=0.44\textwidth, trim={0.8cm 0.8cm 0.2cm 1.5cm},clip]{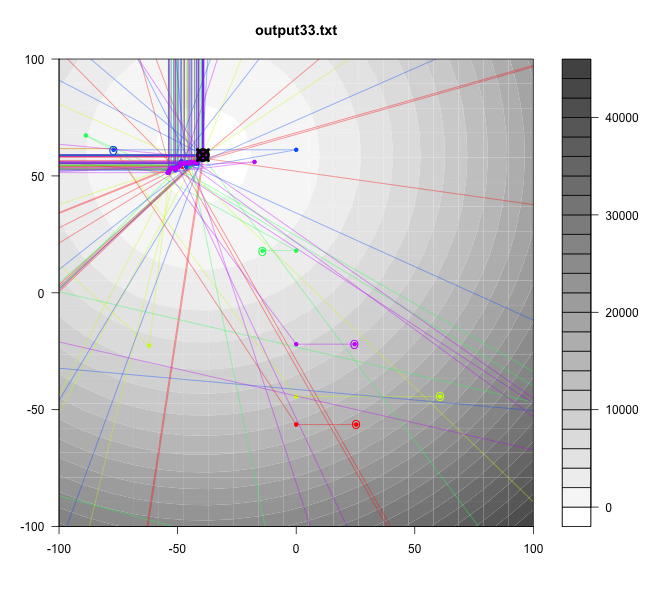}
	\includegraphics[width=0.44\textwidth, trim={0.8cm 0.8cm 0.2cm 1.5cm},clip]{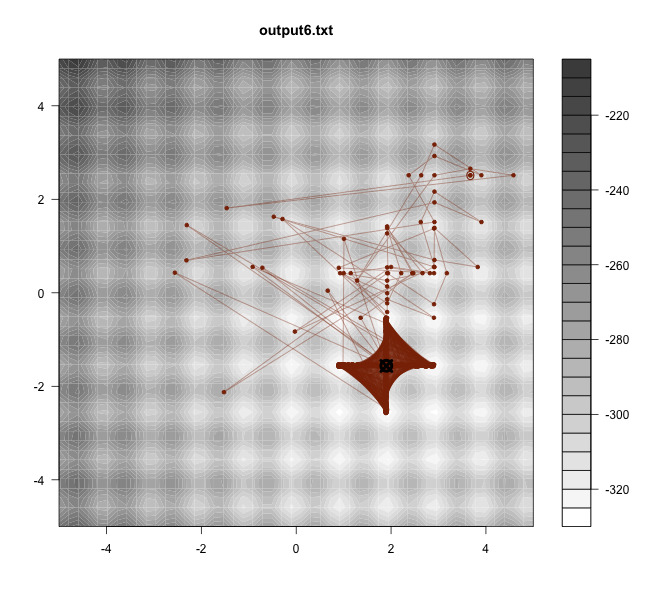}
	\includegraphics[width=0.44\textwidth, trim={0.8cm 0.8cm 0.2cm 1.5cm},clip]{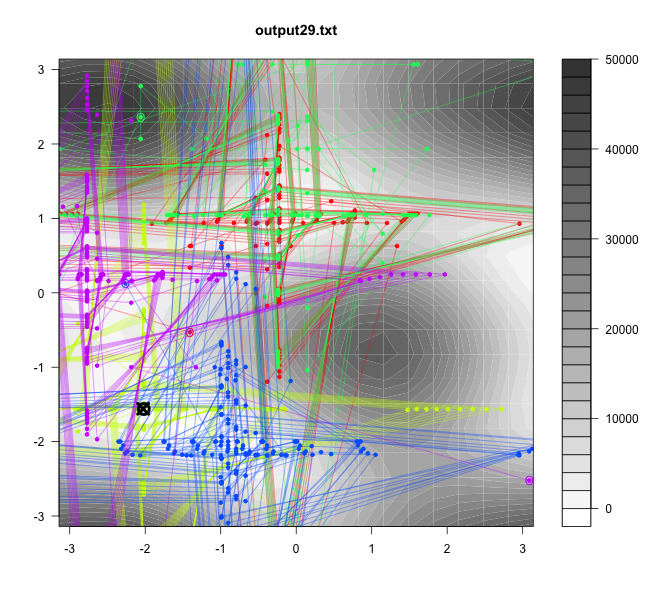}
	\includegraphics[width=0.44\textwidth, trim={0.8cm 0.8cm 0.2cm 1.5cm},clip]{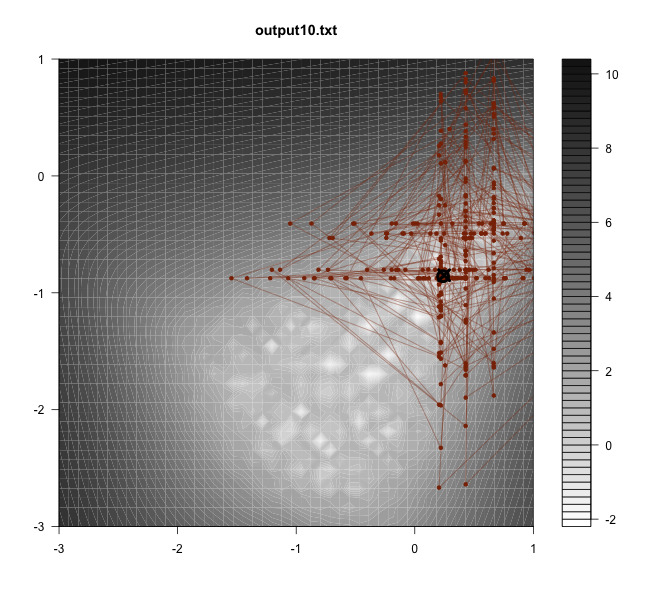}
	\includegraphics[width=0.44\textwidth, trim={0.8cm 0.8cm 0.2cm 1.5cm},clip]{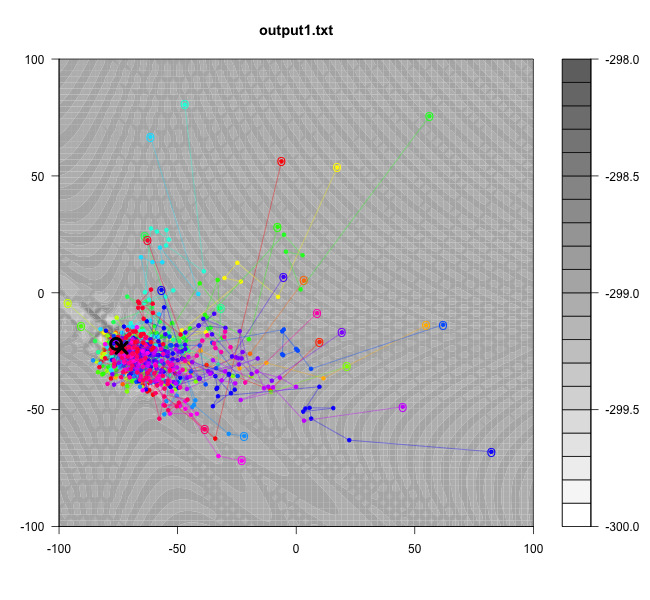}
	\caption{Example trajectories of the best-in-problem optimisers (F1 and F9 top, F12 and F13 middle, F14 bottom) on the 2D versions of the benchmark problems they were trained on. The global minimum is shown as a black circle. The best point reached by the optimiser is shown as a black cross. Each population member's trajectory is shown as a separate colour, with each search point shown as a point. Initial search points are surrounded by small coloured circles. The search landscape is shown in the background as a contour plot.}
	\label{fig:bests}
\end{figure}

Table \ref{tab:optimisers} shows the evolved Push expression used by each best-in-problem optimiser, in each case slightly simplified by removing instructions that have no effect on their fitness. Whilst it is difficult to understand their behaviour by looking at these expressions alone, it is usually possible to gain more insight by observing the interpreter's stack states as they run, and by observing their trajectories on 2D versions of the problems on which they were trained. Fig. \ref{fig:bests} shows examples of the latter; in almost all cases, optimisers generalise well to these easier 2D problems, and it can be seen in each case that the global optimum is found. It can also be seen from the trajectories that the behaviours of the five optimisers are quite diverse, and this is reflected in their program-level behaviours:
\begin{itemize}
	\item Each particle in the $F_1$ optimiser looks up the population best and then adds a random vector to this to generate a new search point. Notably, the size of this random vector is determined using a trigonometric expression based on the components of the particle's current and best search points, meaning that the move size carried out by each particle in the population is different.
	\item The $F_9$ optimiser (which uses only one point of search) continually switches between searching around the best-seen search point and evaluating a random search point. When searching around the best point, at each iteration it adds the sine of the move number to a single dimension, moving along two dimensions each time; in essence, this causes it to systematically explore the nearby search space, building up the space-filling pattern seen in Fig. \ref{fig:bests}.
	\item The $F_{12}$ optimiser is the most complex, and its behaviour at the instruction level is hard to understand. However, it does use the particle's index and the index (but not the vector) of the population best, and both the improvement and out-of-bounds Boolean signals to determine each move. By observing its search trajectories, it is evident that it builds up a geometric pattern that causes it to explore moves with a power series distribution---in essence, a novel form of variable neighbourhood search.
	\item The $F_{13}$ optimiser, by comparison, has the simplest program. Each iteration, it adds a random value to one of the dimensions of the best-seen search point, cycling through the dimensions on each subsequent move (hence why it generates a cross-shaped trajectory). The size of the move (the upper bound of the random value) is determined by both the sine of the objective value of the current point and the sine of the maximum dimension size, the former causing it to vary cyclically as search progresses, and the latter allowing it to adapt the move size to the search area.
	\item The $F_{14}$ optimiser is the only one of these five which uses both a larger population and the \code{vector.between} instruction. Each iteration, it uses this to generate a new population of search points half-way between the population best and one of each particle’s previous positions. Interestingly, which previous position is used for a particular particle is determined by its index; the first particle uses its current position, higher numbered particles go back further in time. This may allow backtracking, which could be useful for landscapes that are deceptive and have limited gradient information (such as $F_{14}$). A small random vector is added to each half-way point, presumably to inject further diversity.
\end{itemize}

\begin{figure}[tb!]
	\centering
	\includegraphics[width=0.19\columnwidth, trim={0.7cm 0 3.5cm 1.5cm},clip]{figs/F1_best.jpg}
	\includegraphics[width=0.19\columnwidth, trim={0.7cm 0 3.5cm 1.5cm},clip]{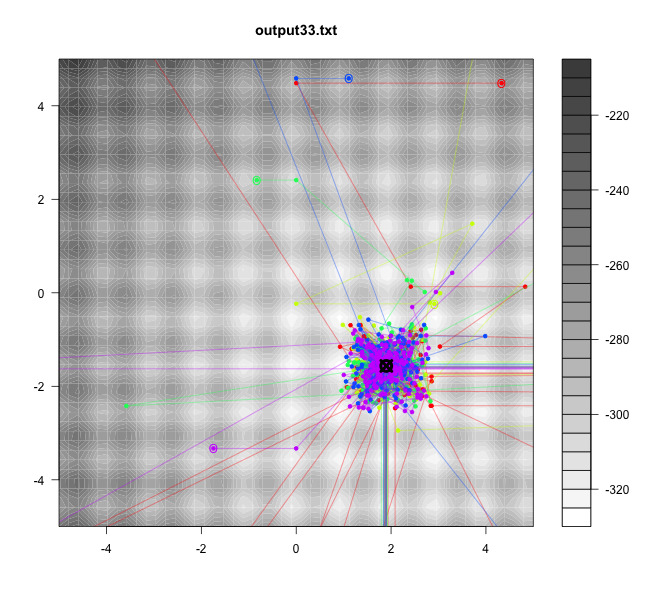}
	\includegraphics[width=0.19\columnwidth, trim={0.7cm 0 3.5cm 1.5cm},clip]{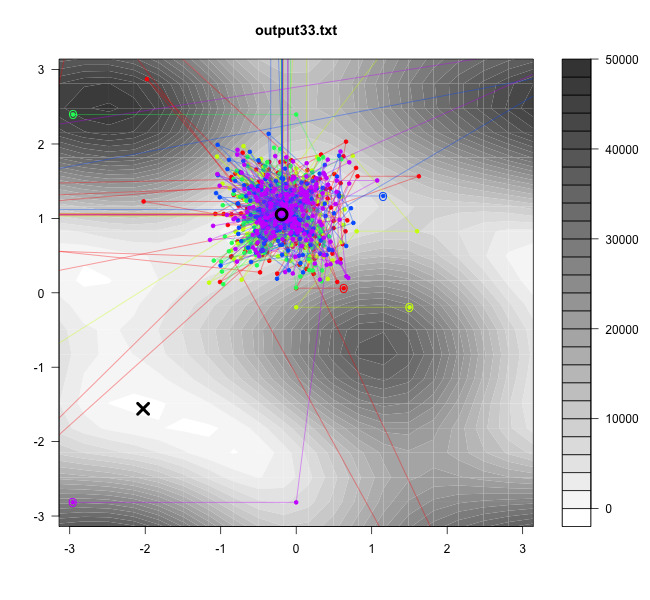}
	\includegraphics[width=0.19\columnwidth, trim={0.7cm 0 3.5cm 1.5cm},clip]{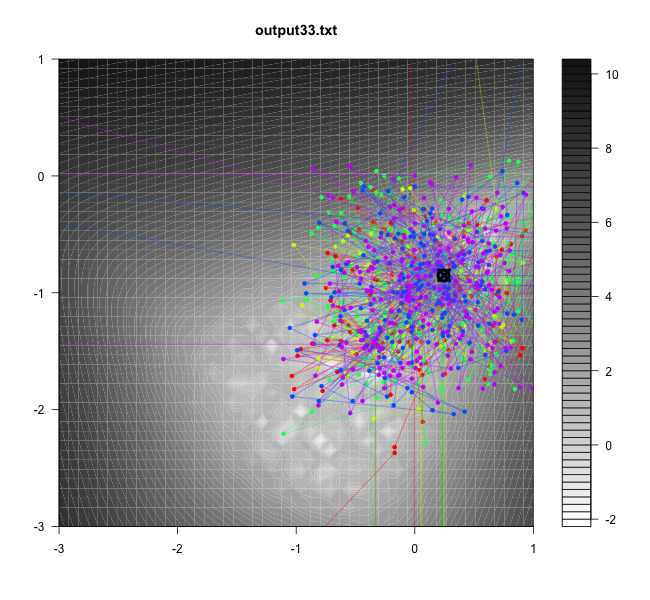}
	\includegraphics[width=0.19\columnwidth, trim={0.7cm 0 3.5cm 1.5cm},clip]{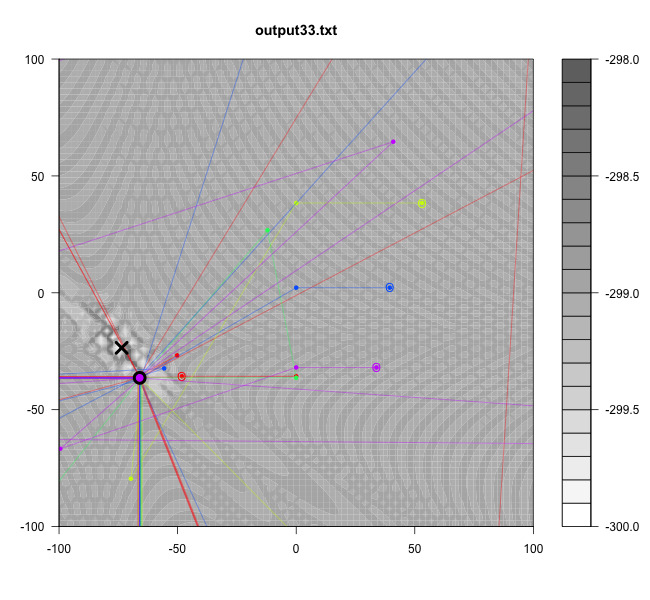}
	
	\includegraphics[width=0.19\columnwidth, trim={0.7cm 0 3.5cm 1.5cm},clip]{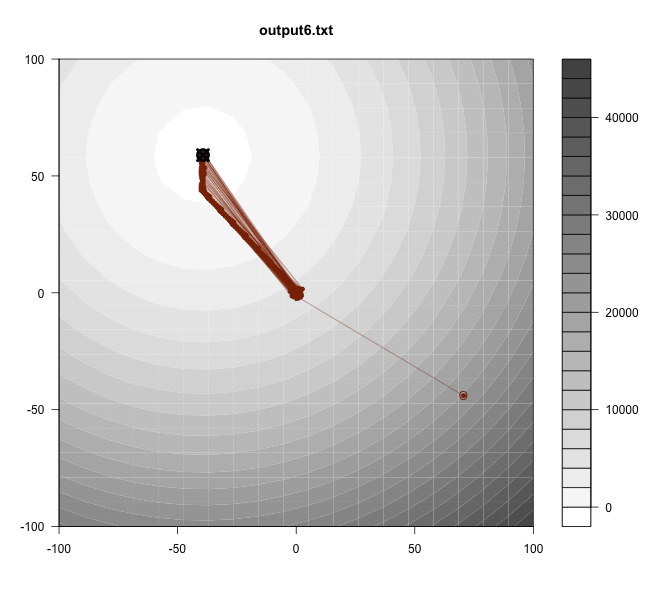}
	\includegraphics[width=0.19\columnwidth, trim={0.7cm 0 3.5cm 1.5cm},clip]{figs/F9_best.jpg}
	\includegraphics[width=0.19\columnwidth, trim={0.7cm 0 3.5cm 1.5cm},clip]{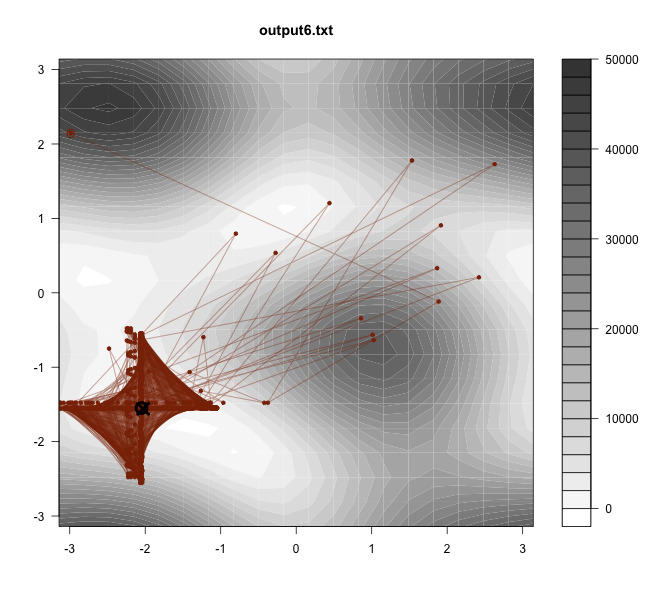}
	\includegraphics[width=0.19\columnwidth, trim={0.7cm 0 3.5cm 1.5cm},clip]{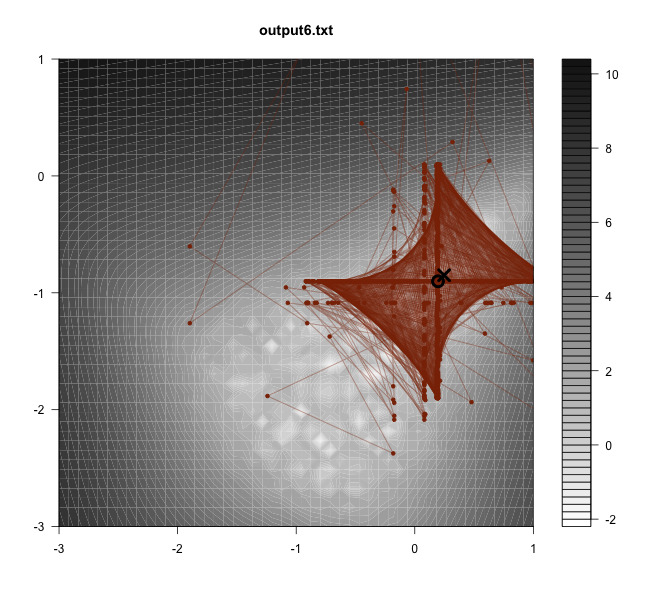}
	\includegraphics[width=0.19\columnwidth, trim={0.7cm 0 3.5cm 1.5cm},clip]{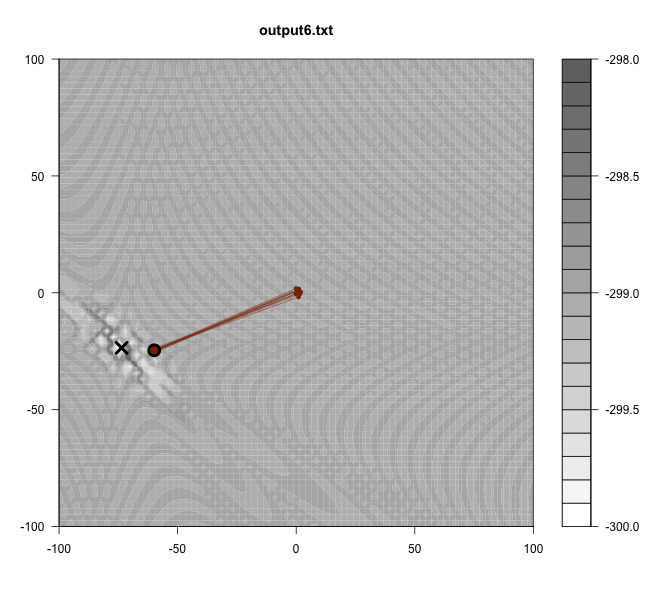}
	
	\includegraphics[width=0.19\columnwidth, trim={0.7cm 0 3.5cm 1.5cm},clip]{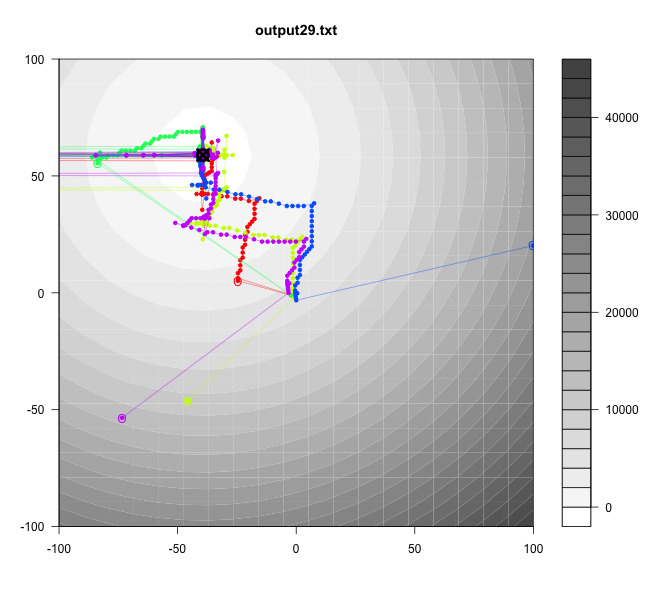}
	\includegraphics[width=0.19\columnwidth, trim={0.7cm 0 3.5cm 1.5cm},clip]{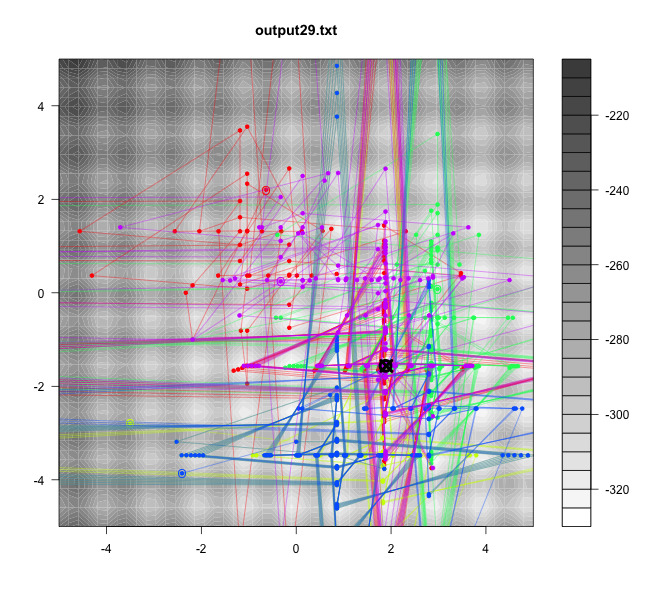}
	\includegraphics[width=0.19\columnwidth, trim={0.7cm 0 3.5cm 1.5cm},clip]{figs/F12_best.jpg}
	\includegraphics[width=0.19\columnwidth, trim={0.7cm 0 3.5cm 1.5cm},clip]{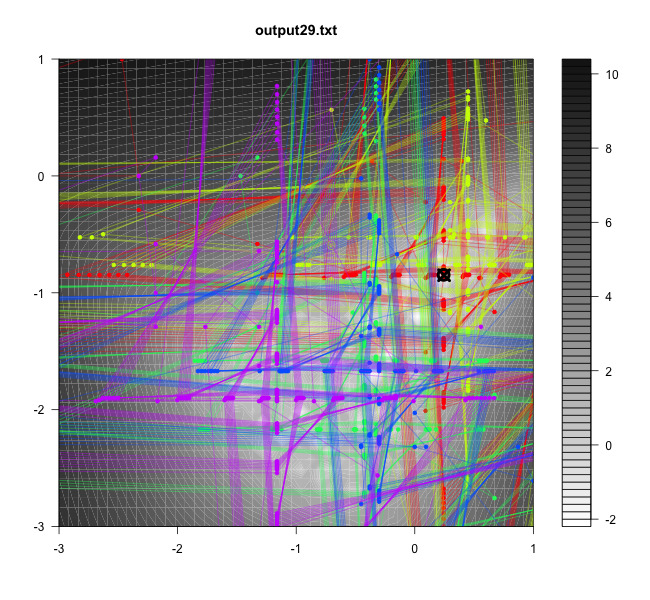}
	\includegraphics[width=0.19\columnwidth, trim={0.7cm 0 3.5cm 1.5cm},clip]{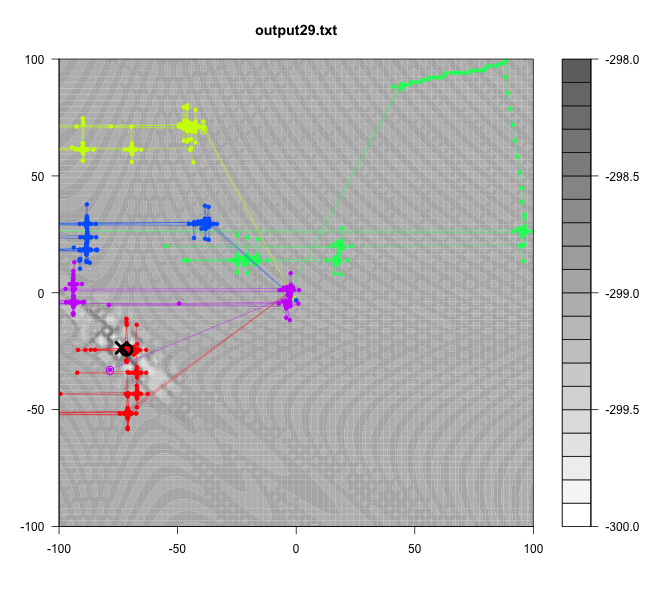}
	
	\includegraphics[width=0.19\columnwidth, trim={0.7cm 0 3.5cm 1.5cm},clip]{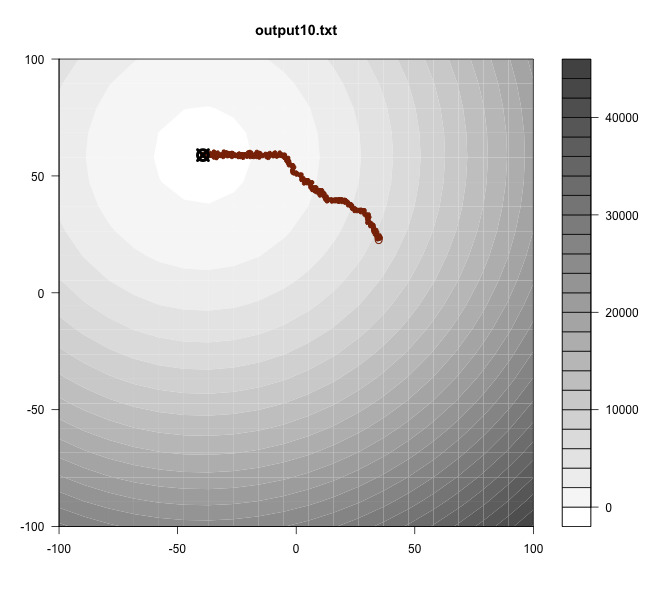}
	\includegraphics[width=0.19\columnwidth, trim={0.7cm 0 3.5cm 1.5cm},clip]{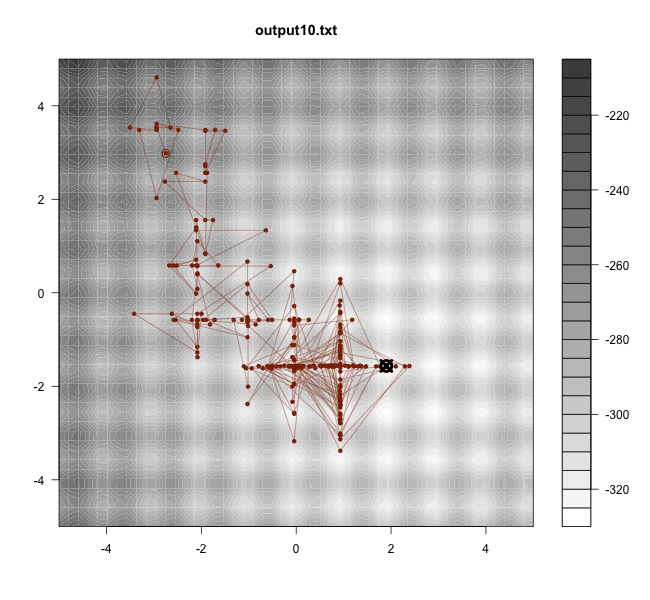}
	\includegraphics[width=0.19\columnwidth, trim={0.7cm 0 3.5cm 1.5cm},clip]{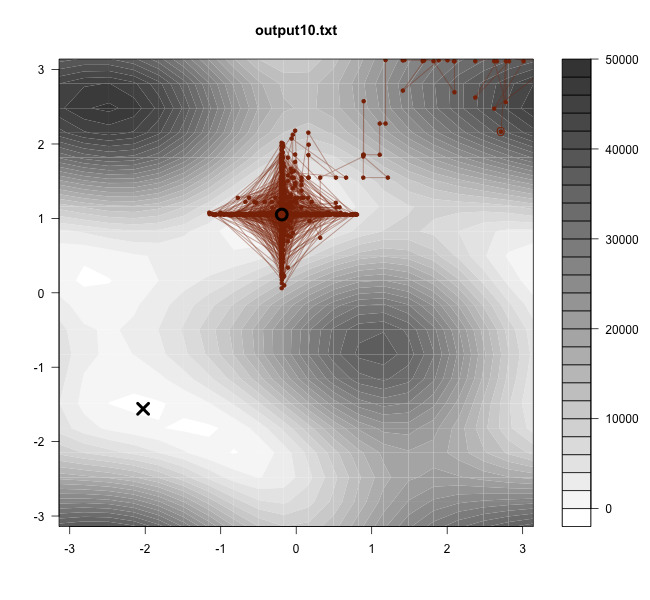}
	\includegraphics[width=0.19\columnwidth, trim={0.7cm 0 3.5cm 1.5cm},clip]{figs/F13_best.jpg}
	\includegraphics[width=0.19\columnwidth, trim={0.7cm 0 3.5cm 1.5cm},clip]{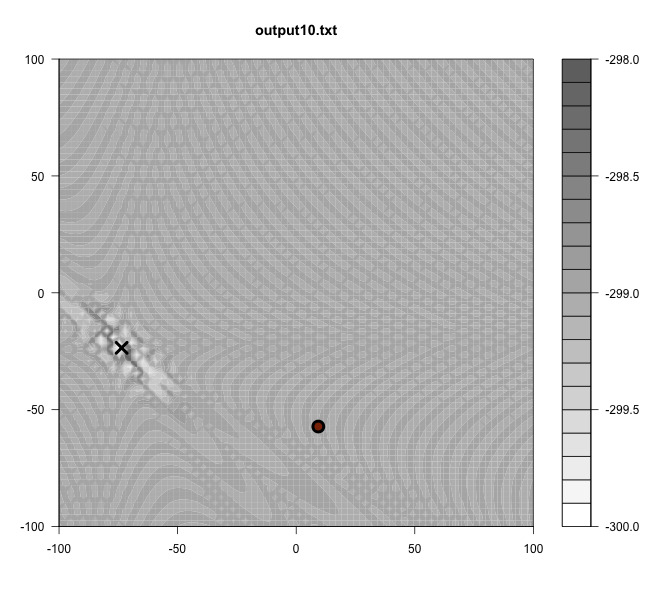}
	
	\includegraphics[width=0.19\columnwidth, trim={0.7cm 0 3.5cm 1.5cm},clip]{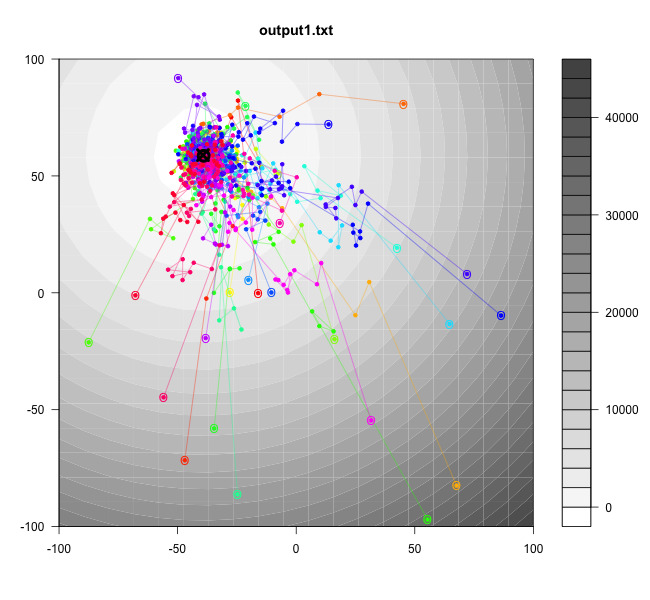}
	\includegraphics[width=0.19\columnwidth, trim={0.7cm 0 3.5cm 1.5cm},clip]{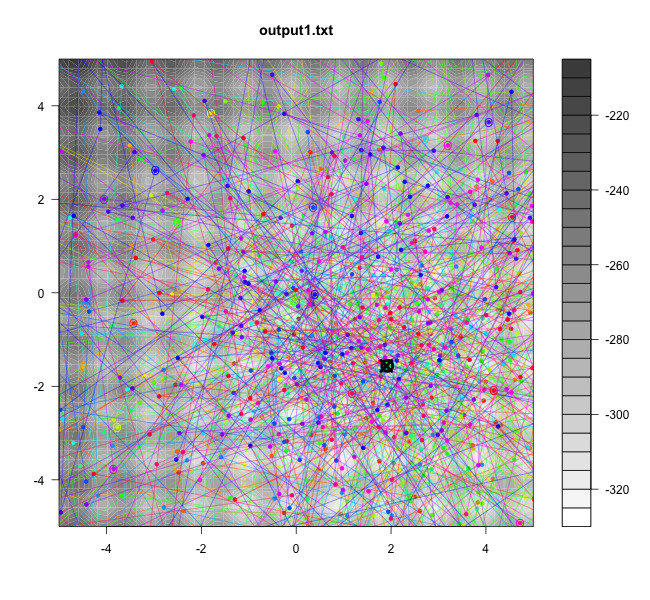}
	\includegraphics[width=0.19\columnwidth, trim={0.7cm 0 3.5cm 1.5cm},clip]{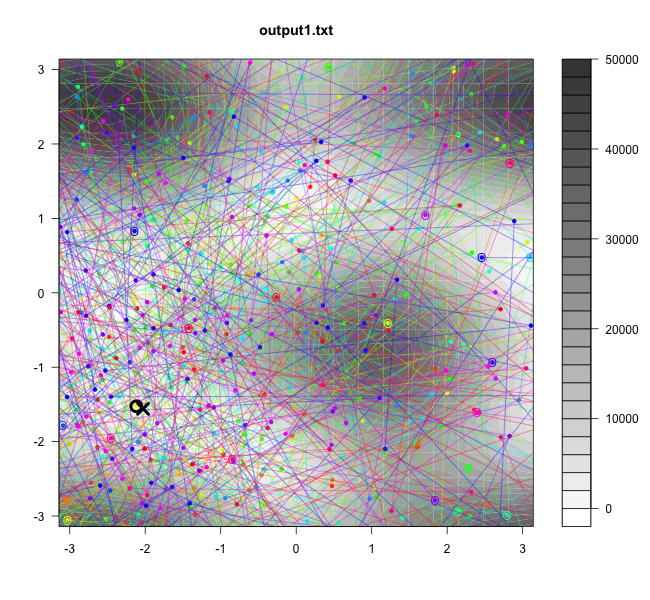}
	\includegraphics[width=0.19\columnwidth, trim={0.7cm 0 3.5cm 1.5cm},clip]{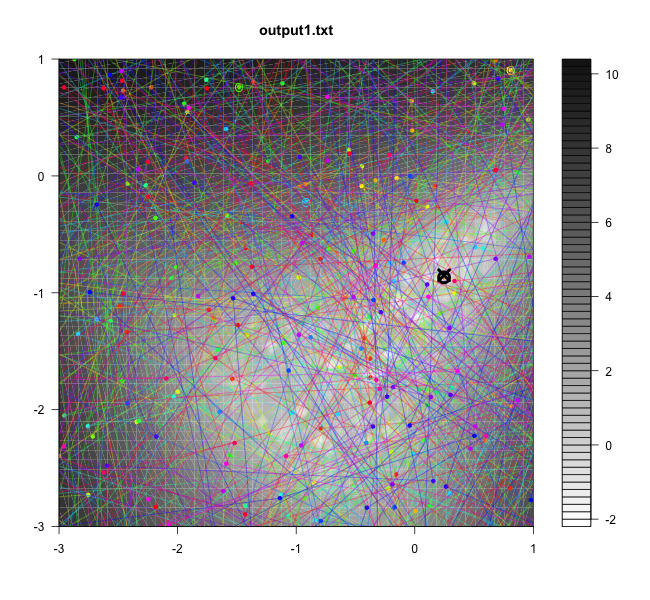}
	\includegraphics[width=0.19\columnwidth, trim={0.7cm 0 3.5cm 1.5cm},clip]{figs/F14_best.jpg}
	\caption{Examples of the best evolved optimisers for each problem (top to bottom: F1,9,12,13,14) applied to each of the other problems (left-right: F1,9,12,13,14). See caption of Figure \ref{fig:bests} for more information.}
	\label{fig:generality}
\end{figure}

Fig. \ref{fig:generality} shows examples of trajectories when each of these optimisers are applied to 2D versions of the other four problems. These suggest that optimisers may fail to generalise not because of intrinsic assumptions about properties of landscapes, but because they make assumptions about the dimensions of the search area. For example, the $F_{9}$ and $F_{13}$ optimisers appear to fail on the $F_{14}$ landscape because they are making moves, or sampling regions, which are only appropriate for a landscape with much smaller overall dimensions. Using a larger range of random scalings during training might help with this.

\begin{figure}[tb!]
	\centering
	\includegraphics[width=0.23\columnwidth, trim={0.7cm 0 3.5cm 1.5cm},clip]{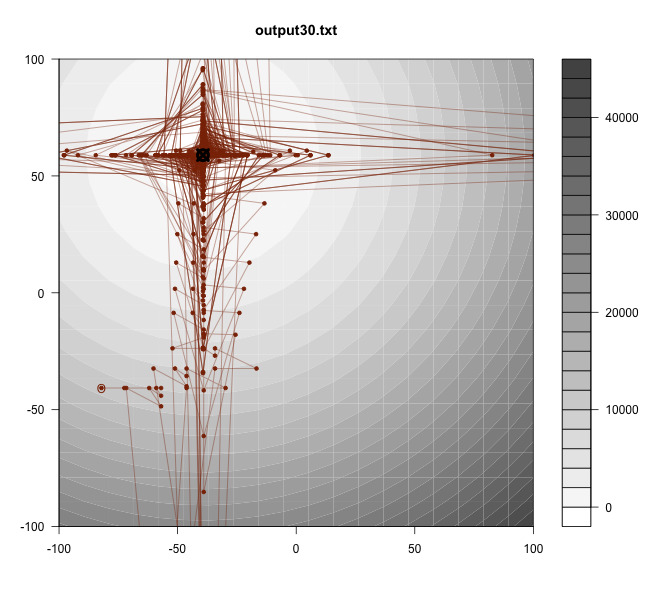}
	\includegraphics[width=0.23\columnwidth, trim={0.7cm 0 3.5cm 1.5cm},clip]{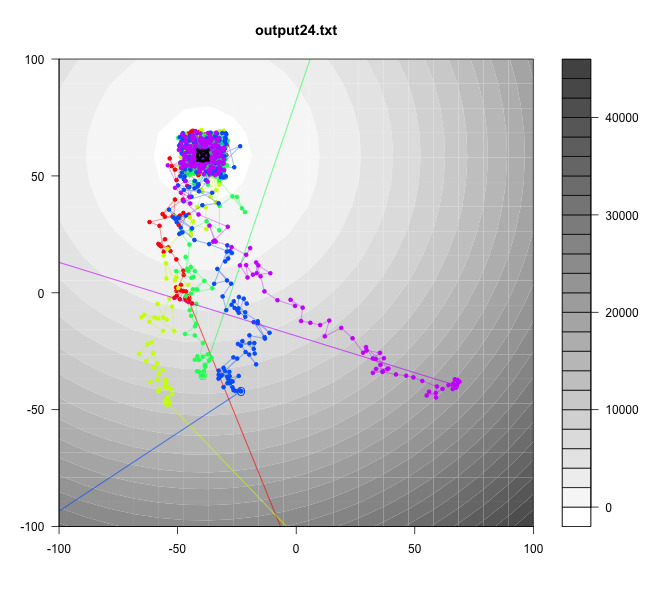}
	\includegraphics[width=0.23\columnwidth, trim={0.7cm 0 3.5cm 1.5cm},clip]{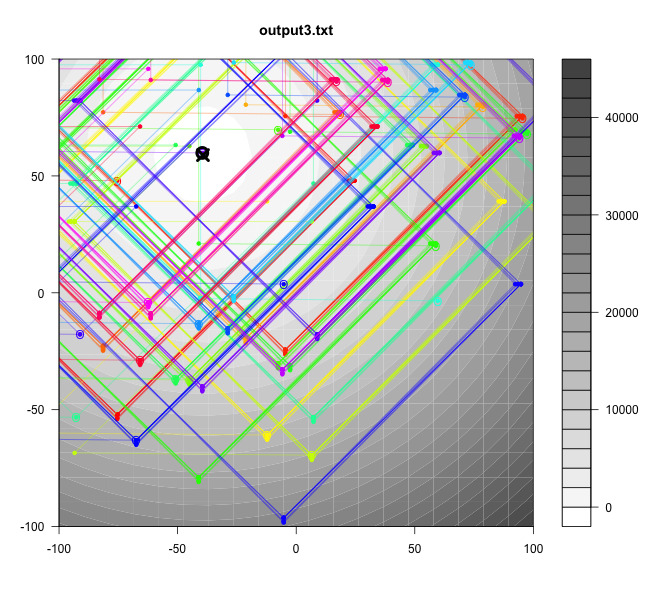}
	\includegraphics[width=0.23\columnwidth, trim={0.7cm 0 3.5cm 1.5cm},clip]{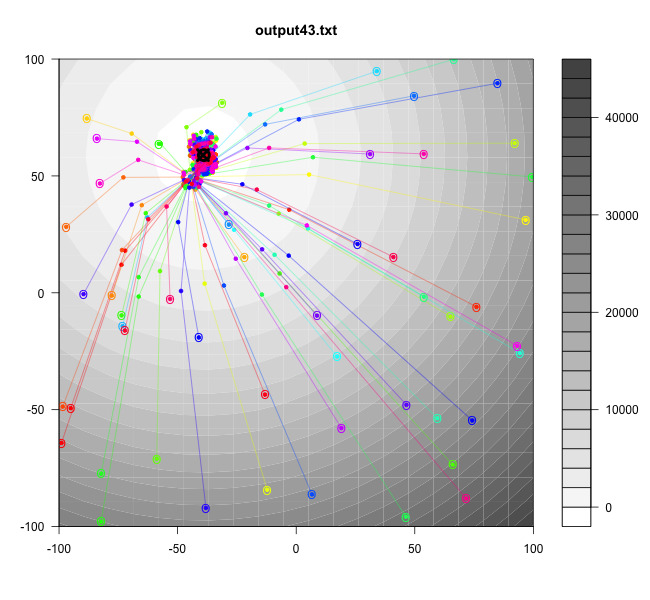}
	
	\includegraphics[width=0.23\columnwidth, trim={0.7cm 0 3.5cm 1.5cm},clip]{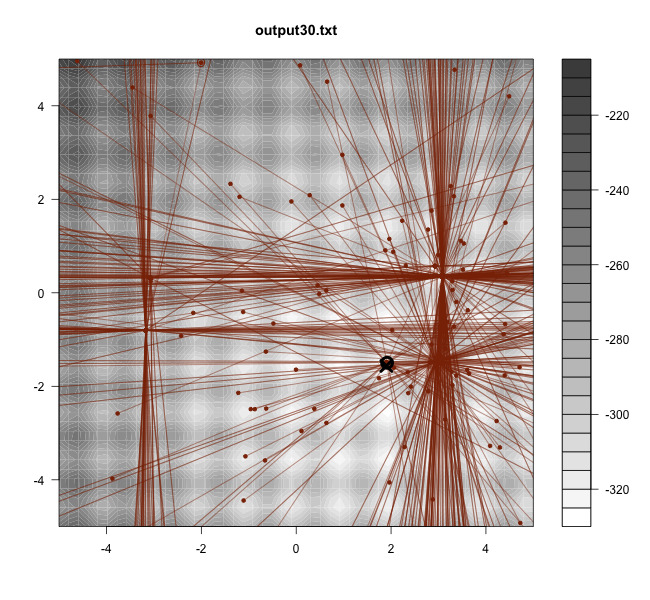}
	\includegraphics[width=0.23\columnwidth, trim={0.7cm 0 3.5cm 1.5cm},clip]{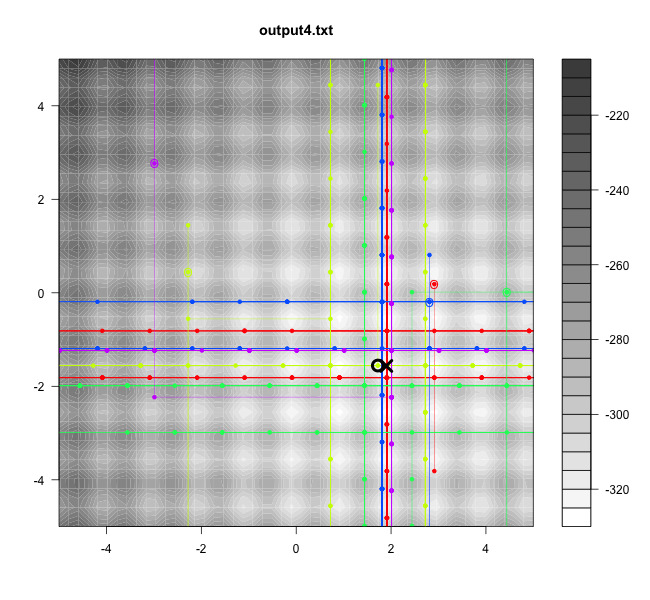}
	\includegraphics[width=0.23\columnwidth, trim={0.7cm 0 3.5cm 1.5cm},clip]{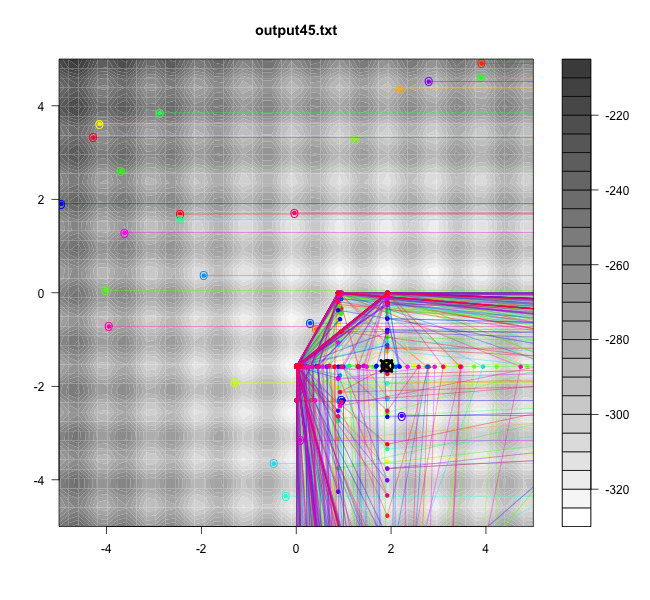}
	\includegraphics[width=0.23\columnwidth, trim={0.7cm 0 3.5cm 1.5cm},clip]{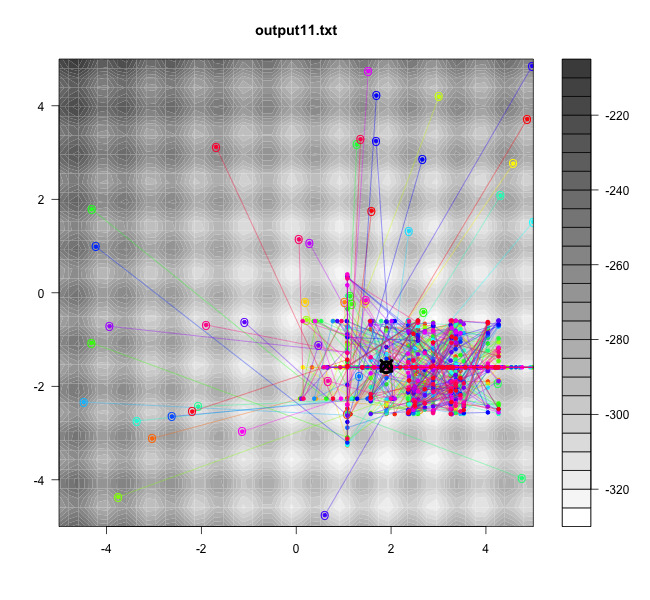}
	
	\includegraphics[width=0.23\columnwidth, trim={0.7cm 0 3.5cm 1.5cm},clip]{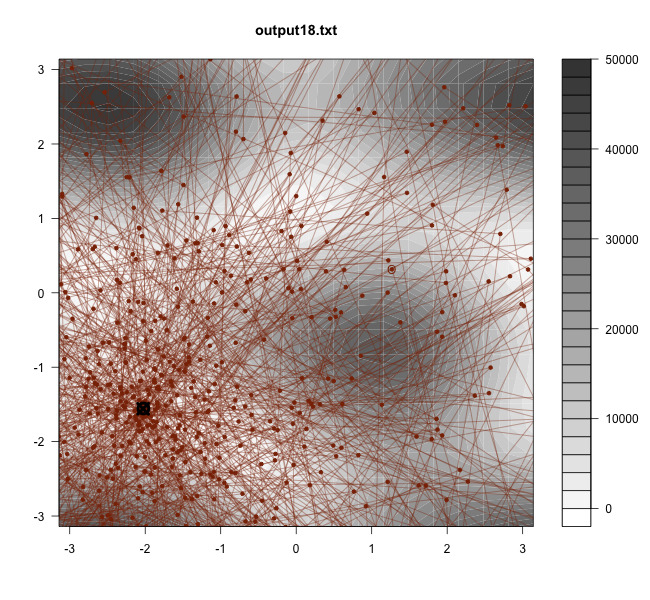}
	\includegraphics[width=0.23\columnwidth, trim={0.7cm 0 3.5cm 1.5cm},clip]{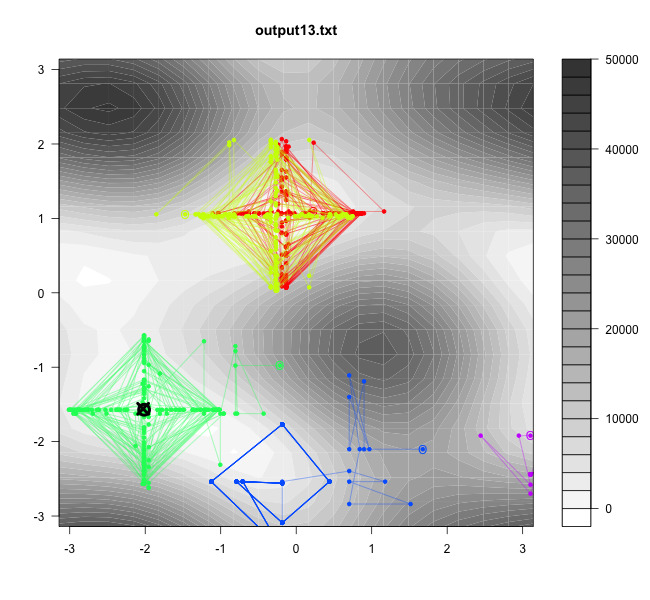}
	\includegraphics[width=0.23\columnwidth, trim={0.7cm 0 3.5cm 1.5cm},clip]{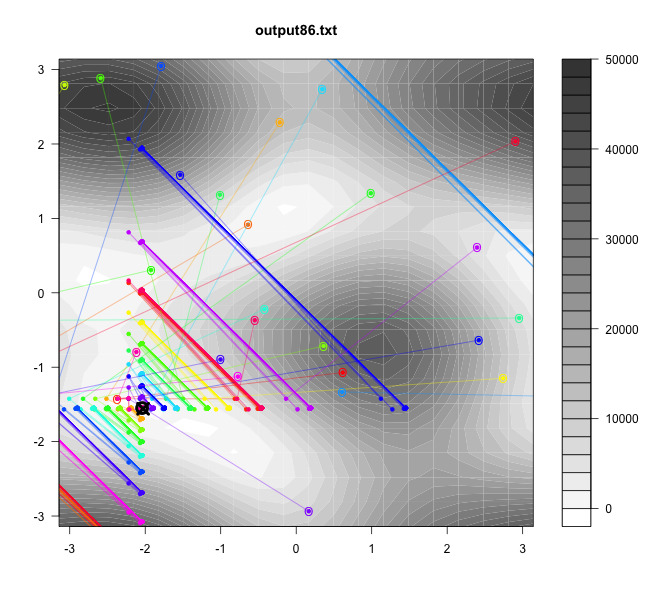}
	\includegraphics[width=0.23\columnwidth, trim={0.7cm 0 3.5cm 1.5cm},clip]{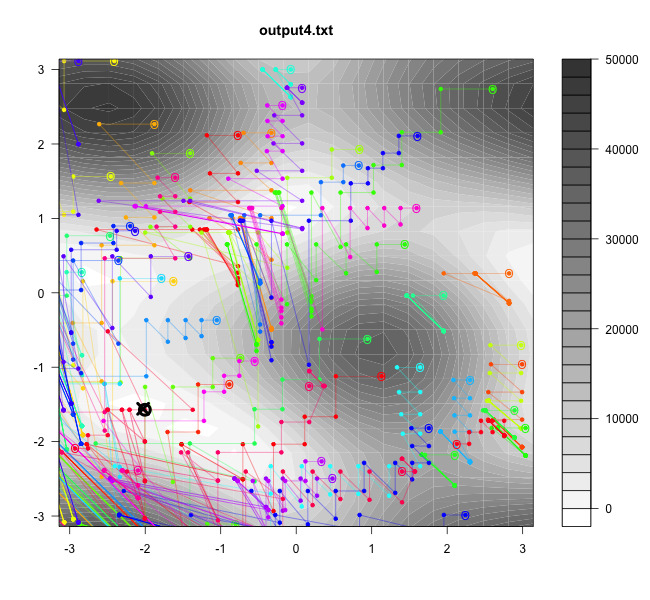}
	
	\includegraphics[width=0.23\columnwidth, trim={0.7cm 0 3.5cm 1.5cm},clip]{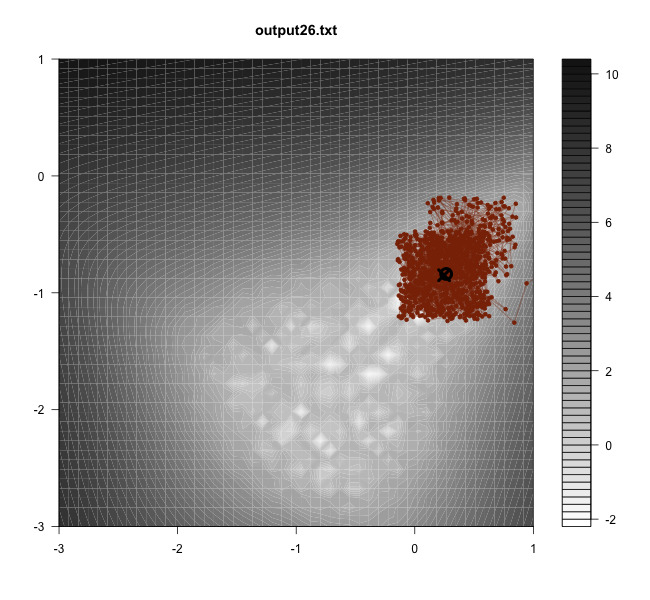}
	\includegraphics[width=0.23\columnwidth, trim={0.7cm 0 3.5cm 1.5cm},clip]{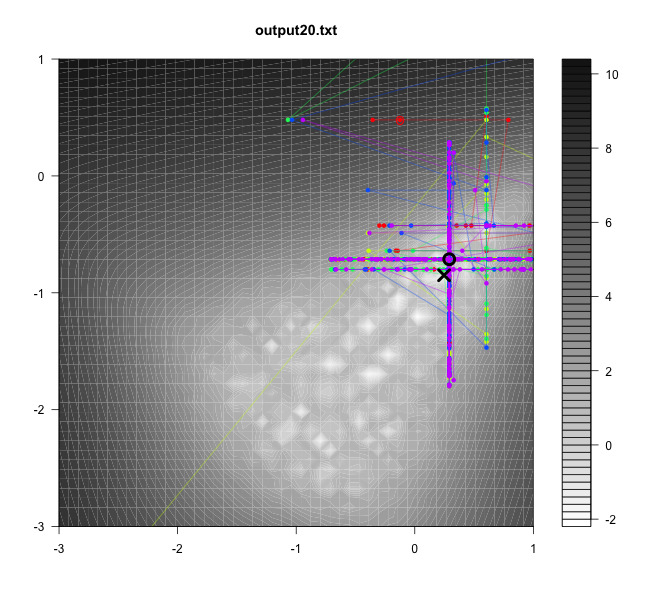}
	\includegraphics[width=0.23\columnwidth, trim={0.7cm 0 3.5cm 1.5cm},clip]{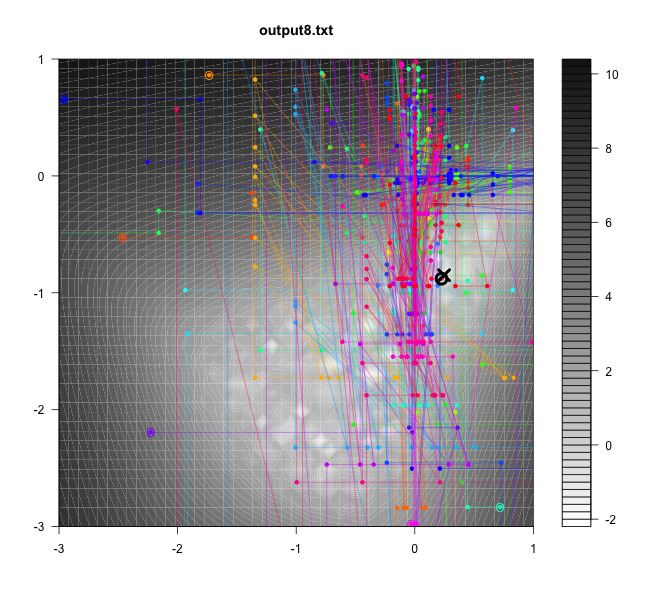}
	\includegraphics[width=0.23\columnwidth, trim={0.7cm 0 3.5cm 1.5cm},clip]{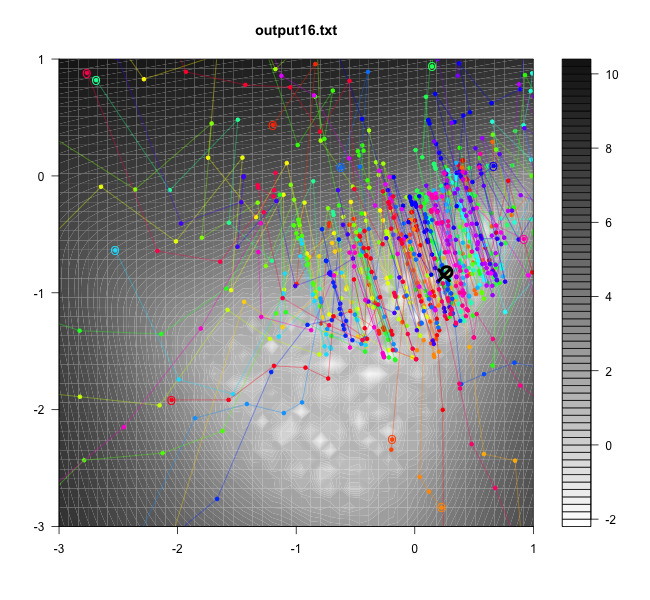}
	
	\includegraphics[width=0.23\columnwidth, trim={0.7cm 0 3.5cm 1.5cm},clip]{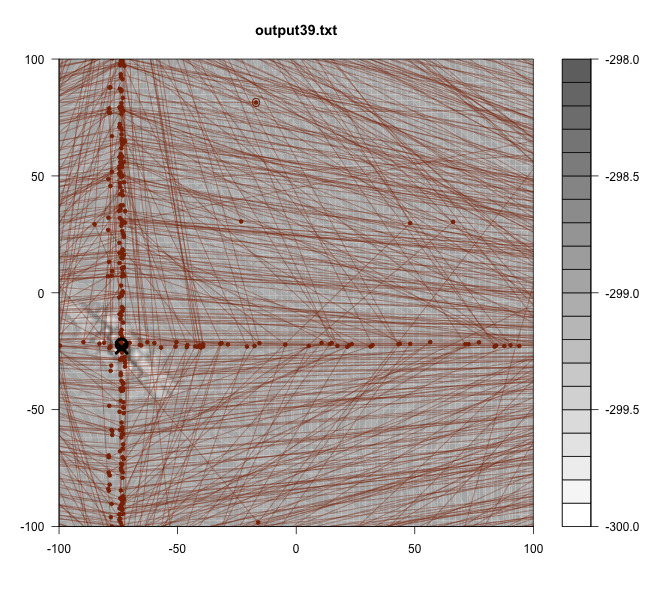}
	\includegraphics[width=0.23\columnwidth, trim={0.7cm 0 3.5cm 1.5cm},clip]{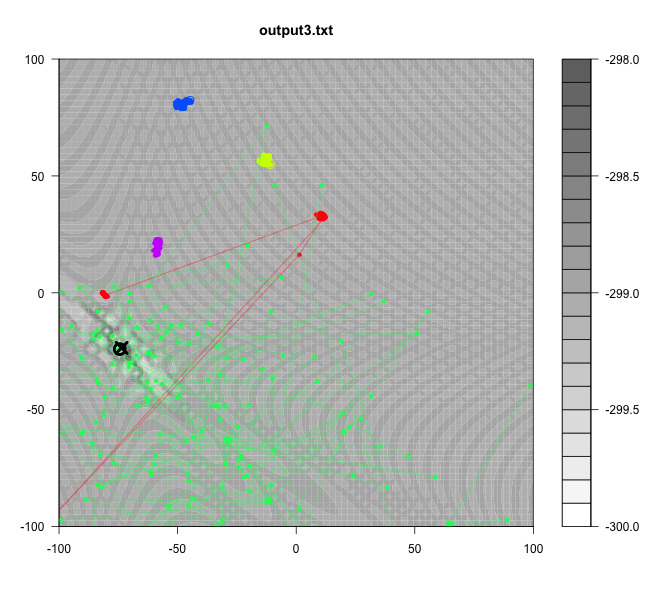}
	\includegraphics[width=0.23\columnwidth, trim={0.7cm 0 3.5cm 1.5cm},clip]{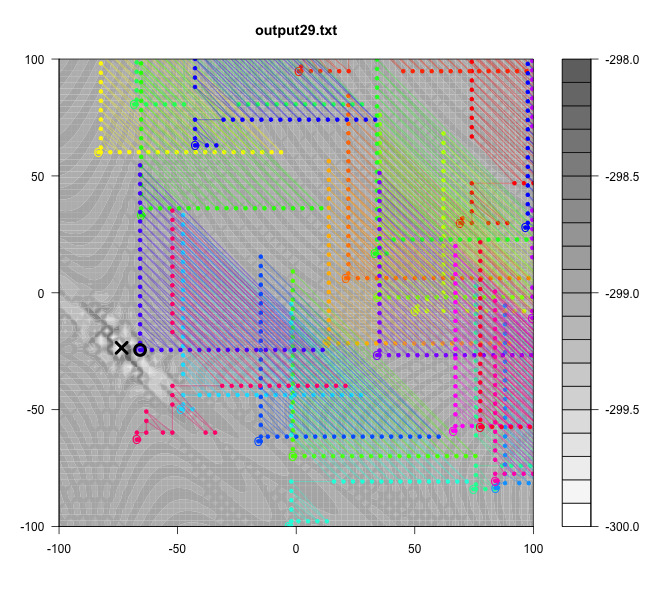}
	\includegraphics[width=0.23\columnwidth, trim={0.7cm 0 3.5cm 1.5cm},clip]{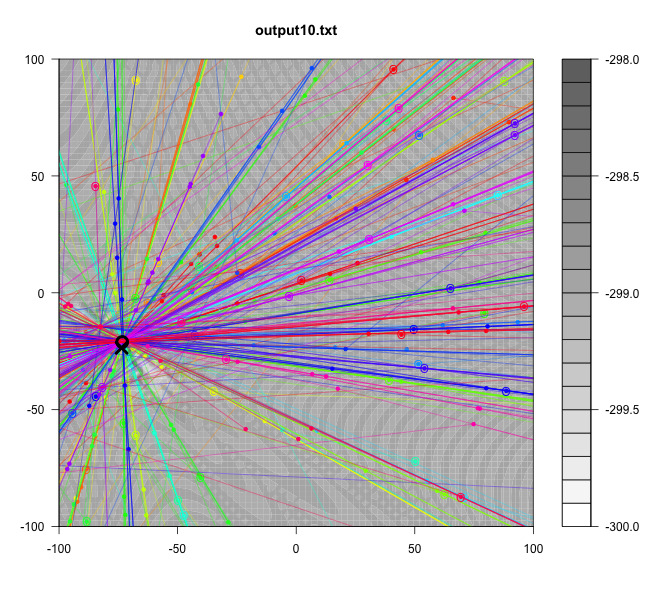}
	\caption{Example trajectories of other evolved optimisers. One example is shown for each combination of problem (top to bottom: F1,9,12,13,14) and population size (left to right: 1, 5, 25, 50). See caption of Figure \ref{fig:bests} for more information.}
	\label{fig:diversity}
\end{figure}

However, these optimisers were not evolved for generality, so the fact that most of them generalise to other problems is a fortunate bi-product. Furthermore, it is likely that the optimisers that do best on one problem are not likely to be the best in terms of generality. Hence, in practice  there is likely to be a benefit to looking at the best optimisers from the other 245 runs depicted in Fig. \ref{fig:distributions}. Fig. \ref{fig:diversity} gives a snapshot of these, showing one example for each combination of training problem and optimiser population size. This illustrates some of the broad diversity seen amongst the solutions. Many of these trajectories look nothing like conventional optimisers, so it is likely that interesting ideas of how to do optimisation could be gained by looking more closely at them. Another interesting direction for future work would be to consider ensembles of optimisers. There are many potential ways of doing this. For example, early results suggest that it may be advantageous, in terms of generality, to form a heterogenous population-based optimiser by combining the best programs from multiple runs.

\section{Conclusions} \label{conclusions}
In recent years, there has been a lot of criticism of the \textit{ad hoc} design of new optimisers through mimicry of natural phenomena. Despite early success with evolutionary algorithms and particle swarm optimisation, this trend has increasingly resulted in optimisers that are technically novel, but which differ in minor and often arbitrary ways from existing optimisers.  If we are to create new optimisation algorithms (and the no free lunch theorem \citep{wolpert1997no} suggests a need for diverse optimisers), then perhaps it is better to do this in a more systematic, objective and automated manner. This paper contributes towards this direction of research by investigating the utility of Push GP for exploring the space of optimisers. The results show that Push GP can both discover and express optimisation behaviours that are effective, complex and diverse. Encouragingly, the evolved optimisers scale to problems they did not see during training, and often out-perform general purpose optimisers on these previously unseen problems. The behavioural analysis shows that the evolved optimisers use a diverse range of metaheuristic strategies to explore optimisation landscapes, using behaviours that differ significantly from existing local and population-based optimisers. Furthermore, these are only the tip of the iceberg; the evolved optimiser populations appear to contain broad behavioural diversity, and there are many potential ways of combining diverse optimisers to create ensembles.

%-----------------------------------------
\bibliographystyle{abbrvnat}
\bibliography{lones}
%-----------------------------------------

\end{document}